\documentclass{article}
\usepackage{ijcai25}

\usepackage{times}
\usepackage{soul}
\usepackage{url}
\usepackage[hidelinks]{hyperref}
\usepackage[utf8]{inputenc}
\usepackage[small]{caption}
\usepackage{graphicx}
\usepackage{amsmath}
\usepackage{amsthm}
\usepackage{booktabs}
\usepackage{algorithm}
\usepackage{algorithmic}
\usepackage[switch]{lineno}

\usepackage{wrapfig}
\usepackage{stmaryrd}
\usepackage{amsfonts}
\usepackage{nicefrac}
\usepackage{microtype}
\usepackage{xcolor}         
\usepackage{tcolorbox}
\usepackage{mdframed}
\usepackage{empheq}
\usepackage{listings}
\tcbuselibrary{breakable}
\usepackage{tikz}

\newcommand*\GOR{\ |\ }
\newcommand{\name}{\textsc{InnateCoder}}
\newtheorem{definition}{Definition}
\newcommand{\modtext}[1]{\textcolor{black}{#1}}

\title{InnateCoder: Learning Programmatic Options with Foundation Models}

\author{
Rubens O. Moraes$^1$
\and
Quazi Asif Sadmine$^{2,3}$\and
Hendrik Baier$^{4,5}$\And
Levi H. S. Lelis$^{2, 3}$\\
\affiliations
$^1$Departamento de Informática, Universidade Federal de Viçosa\\
$^2$Department of Computing Science, University of Alberta \\
$^3$Alberta Machine Intelligence Institute (Amii) \\
$^4$Information Systems, Eindhoven University of Technology\\
$^5$Centrum Wiskunde \& Informatica, Amsterdam\\
}

\begin{document}

\maketitle

\begin{abstract}
Outside of transfer learning settings, reinforcement learning agents start their learning process from a clean slate. As a result, such agents have to go through a slow process to learn even the most obvious skills required to solve a problem. In this paper, we present \name, a system that leverages human knowledge encoded in foundation models to provide programmatic policies that encode ``innate skills'' in the form of temporally extended actions, or options. In contrast to existing approaches to learning options, \name\ learns them from the general human knowledge encoded in foundation models in a zero-shot setting, and not from the knowledge the agent gains by interacting with the environment. Then, \name\ searches for a programmatic policy by combining the programs encoding these options into larger and more complex programs. We hypothesized that \name's way of learning and using options could improve the sampling efficiency of current methods for learning programmatic policies. Empirical results in MicroRTS and Karel the Robot support our hypothesis, since they show that \name\ is more sample efficient than versions of the system that do not use options or learn them from experience. 
\end{abstract}

\section{Introduction}

Deep reinforcement learning (DRL) agents typically begin their training process with randomly initialized neural networks. As a result, they must learn from scratch even the most basic skills required to solve a problem. Due to the high sample complexity of DRL algorithms, such \emph{tabula rasa} learning can be inefficient and time-consuming, which has inspired several research directions aiming at reusing prior computation~\cite{transferrl,continualrl}.

In this paper, we harness the general human knowledge encoded in foundation models to endow agents with helpful skills before they even start interacting with the environment. This is achieved by using programmatic representations of policies~\cite{leaps}---programs written in a domain-specific language encoding policies---and the foundation models' ability to write computer programs. Depending on the language used, programmatic policies can generalize better to unseen scenarios and be human-interpretable~\cite{VermaMSKC18,viper}. 
In addition to these advantages and loosely inspired by the innate abilities of animals~\cite{Tinbergen1951}, we show that these representations of policies allow us to harness helpful ``innate skills'' from foundation models.

Given a natural-language description of the problem that the agent needs to learn to solve, our system, which we call \name, queries a foundation model for programs that encode policies to solve the problem. 
Although the programs the model generates are unlikely to encode policies that solve the problem, we hypothesize that the set of sub-programs we obtain from these programs can encode helpful temporally extended actions, or options~\cite{options_sutton}. 
We consider options as functions the agent can call and that will tell it how to act for a number of steps~\cite{precup_options}. 

Options can improve the agent's learning in different ways. They can allow the agent to better explore the problem space~\cite{machado2017laplacian} or transfer knowledge between different tasks~\cite{konidaris2007building}. This paper presents a novel way of learning options with foundation models. We also present a novel way of using the learned options, which is inspired by recent work on learning semantic spaces of programming languages~\cite{semantic_spaces}. 
In the semantic space, neighbor programs encode similar but different agent behavior, which can be conducive to algorithms searching for programmatic policies~\cite{leaps}. 
Instead of searching only in the space of programs induced by the syntax of the language, \name\ also searches in the semantic space induced by options. While previous methods can benefit from up to hundreds of options~\cite{eysenbach2019diversity}, \name's use of programmatic options allows it to benefit from thousands of them. 

\name's approach to harnessing options from foundation models contrasts with previous approaches to automatically learning them, e.g., \cite{tessler2017deep,bacon2017option,igl2020multitask,klissarov2023deep}. This is because options are harnessed from the general knowledge encoded in a foundation model, as opposed to the knowledge the agent gains by interacting with the environment. This zero-shot approach to learning options is enabled by the use of a domain-specific language to bridge the gap between the high-level knowledge encoded in foundation models and the low-level knowledge required at the sensorimotor control level of the agent~\cite{klissarov2024motif}. 
For example, foundation models trained on Internet data likely encode the knowledge that, to win a match of a real-time strategy game, the player must collect resources and build structures, which will allow for the training of the units needed to win the game. However, the model cannot issue low-level actions in real time to control dozens of units to accomplish this plan. \name\ helps bridge this gap by distilling the knowledge of the model into options that can be executed in real time. 

We evaluated our hypothesis that foundation models can generate helpful programmatic options in the domains of MicroRTS, a challenging real-time strategy game~\cite{ontanon2017combinatorial}, and Karel the Robot~\cite{karel1994}, a benchmark for program synthesis and reinforcement learning algorithms~\cite{executionguided2018,leaps}. The results in both domains support our hypothesis, since \name\ was more sample-efficient than versions of the system that do not use options or learn options from experience. We also show that the policies \name\ learns are competitive and often outperform the current state-of-the-art algorithms. \name\ is inexpensive because it uses the foundation model a small number of times as a pre-processing step, making it an accessible system to smaller labs and companies.\footnote{\name\ is available at \url{https://github.com/rubensolv/InnateCoder}.}  

\section{Problem Definition}

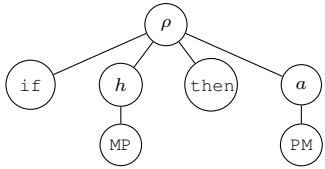
\begin{figure}[t]
    \centering
    \begin{minipage}[c]{0.3\textwidth} 
        \begin{align*}
            \rho :=~& \texttt{ if }h \texttt{ then } a \\
            h :=~& \texttt{frontIsClear }|\texttt{ markersPresent } \\
            a :=~& \texttt{move }| \texttt{ putMarker }|\texttt{ pickMarker }
        \end{align*}
    \end{minipage}%
    \hspace{0.05\textwidth} 
    \begin{minipage}[c]{0.4\textwidth} 
    \hspace{0.5cm}
    \begin{tikzpicture}[
            level distance=8mm,
            every node/.style={text width=0.3cm,align=center,circle,draw,},
            level 1/.style={sibling distance=12mm},
            level 2/.style={sibling distance=12mm},
            level 3/.style={sibling distance=9mm}
        ]
        \scriptsize
        \node {\texttt{$\rho$}}
                child {node [inner sep=0.5mm, text width=0.5cm] {\texttt{if}}}
                child {node {$h$}
                    child {node {\texttt{MP}}}
                }
                child {node [inner sep=0.5mm, text width=0.55cm] {\texttt{then}}}
                child {node {$a$}
                        child {node [inner sep=0.15mm, text width=0.5cm] {\texttt{PM}}}
                };
        \end{tikzpicture}
    \end{minipage}
    \caption{\modtext{Top: Context-free grammar specifying a simplified version of the domain-specific language for Karel the Robot. Bottom: Abstract syntax tree for \texttt{if markersPresent then pickMarker}, where \texttt{MP} and \texttt{PM} stand for \texttt{markersPresent} and \texttt{pickMarker}, respectively. Karel is a robot acting on a grid, where it needs to accomplish tasks such as collecting and placing markers on different locations of the grid. In this program, Karel will pick up a marker if one is present in its current location on the grid.}}
    \label{fig:dsl}
\end{figure}

We consider sequential decision-making problems formulated as Markov decision processes (MDPs) $(S, A, p, r, \mu, \gamma)$. Here, $S$ is the set of states and $A$ the set of actions. The function $p(s_{t+1}|s_t, a_t)$ is the transition model, which gives the probability of reaching state $s_{t+1}$ given that the agent is in $s_t$ and takes action $a_t$ at time step $t$. The agent observes a reward value of $R_{t+1}$ when transitioning from $s_t$ to $s_{t+1}$. The reward value the agent observes is returned by the function $r$. $\mu$ is the distribution of the initial states of the MDP; states sampled from $\mu$ are denoted $s_0$. $\gamma$ in $[0, 1]$ is the discount factor. A policy $\pi$ is a function that receives a state $s$ and returns a probability distribution over actions available at $s$. The goal is to learn a policy $\pi$ that maximizes the expected sum of discounted rewards for $\pi$ starting in an $s_0$: $\mathbb{E}_{\pi,p,\mu}[\sum_{k=0}^\infty \gamma^k R_{k+1}]$. $V^\pi(s) = \mathbb{E}_{p, \pi}[\sum_{k=0}^\infty \gamma^k R_{k+1} | s_0 = s]$ is the value function that measures the expected return for policy $\pi$ starting from $s$. In this work, we approximate the value function of a policy $\pi$ and state $s$ with Monte Carlo roll-outs and denote the approximation as $\hat{V}^\pi(s)$.

We consider programmatic representations of policies, which are policies written in a domain-specific language (DSL). The set of programs a DSL accepts is defined through a context-free grammar $(M, N, R, I)$, where $M$, $N$, $R$, and $I$ are the sets of non-terminals, terminals,
the production rules, and the grammar's initial symbol, respectively. Figure~\ref{fig:dsl} shows a DSL for a simplified version of the language we use for Karel the Robot (the complete DSL is shown in Appendix~\ref{sec:karel_dsl}). In this DSL, the set $M$ is composed of symbols $\rho$, $h$, $a$, while the set $N$ includes the symbols \texttt{if},  \texttt{frontIsClear}, \texttt{markersPresent}, \texttt{move}, \texttt{putMarker}, \texttt{pickMarker}. $R$ are the production rules (e.g., $h \to \texttt{frontIsClear}$), and $\rho$ is the initial symbol. We denote programmatic policies with letters $p$ and $n$ and their variations such as $n'$ and $n^*$. 

We represent programs as abstract syntax trees (AST), where each node $n$ and its children represent a production rule if $n$ represents a non-terminal symbol. For example, the root of the tree in Figure~\ref{fig:dsl} represents the non-terminal $\rho$, while node $\rho$ and its children represent the production rule $\rho \to \texttt{if } h \texttt{ then } a$. Leaf nodes in the AST represent terminal symbols. 
Figure~\ref{fig:dsl} shows an example of an AST for the program \texttt{if markersPresent then pickMarker}. A DSL $D$ defines the possibly infinite space of programs $\llbracket D \rrbracket$, where in our case each program $p$ in $\llbracket D \rrbracket$ represents a policy.

Given a domain-specific language $D$, our task is to find a programmatic policy $p \in \llbracket D \rrbracket$ that maximizes the expected sum of discounted rewards for a given MDP.

\begin{figure*}[t]
    \centering
\includegraphics[width=0.7\textwidth]{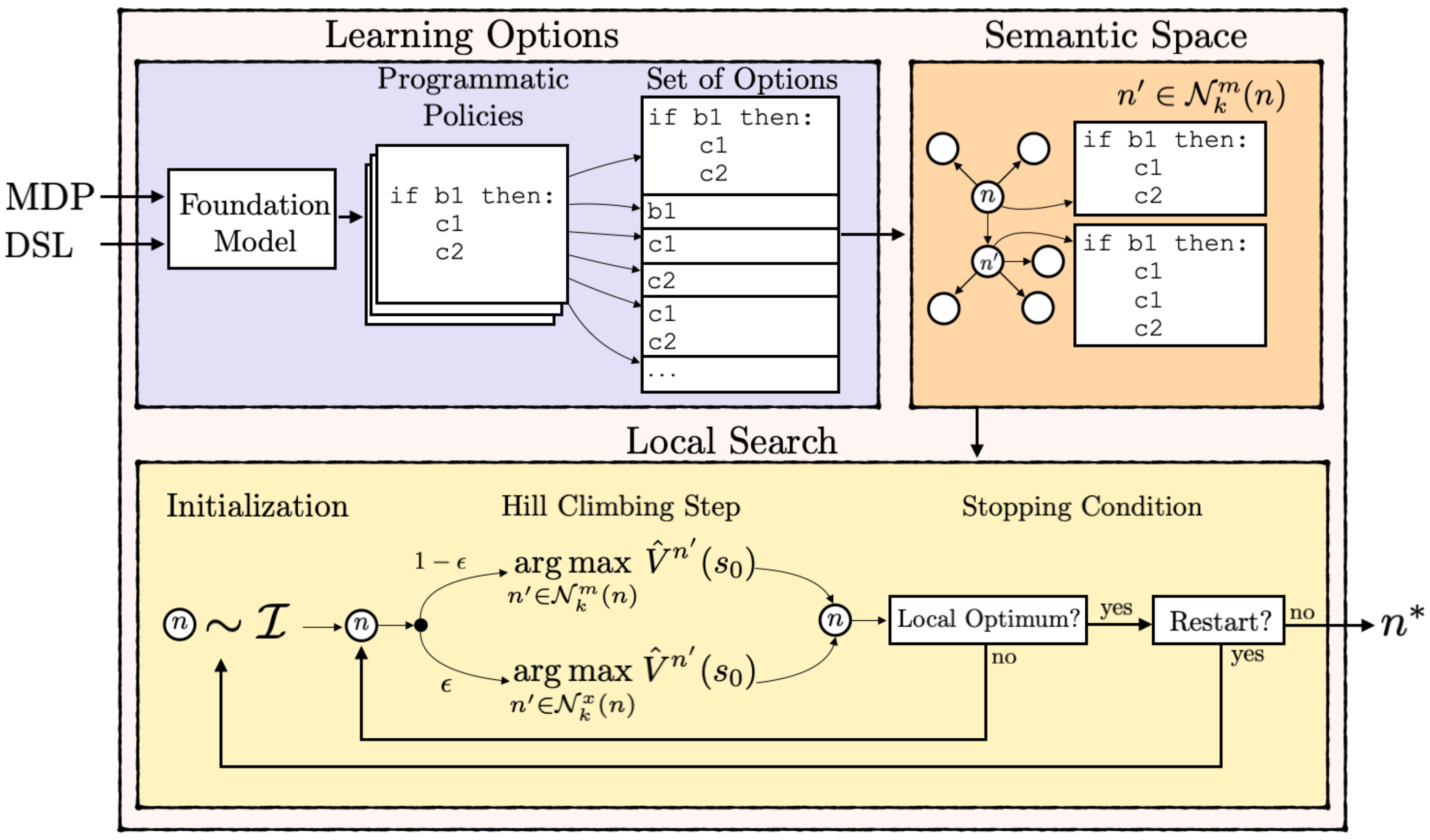}
    \caption{\name\ has three parts. ``Learning Options'' harnesses options from a foundation model. The model generates a set of programmatic policies that are broken into a set of options (Section~\ref{sec:options}). ``Semantic Space'' uses the options to induce the semantic space of the DSL  (Section~\ref{sec:semantic}). ``Local Search'' searches in a mixture of the syntax and semantic spaces for a policy $n^*$ (Section~\ref{sec:searching}).}
    \label{fig:innate_coder}
\end{figure*}

\section{\name}

Figure~\ref{fig:innate_coder} shows a schematic view of \name with its three components: one for learning options, another that uses the options to induce an approximation of the semantic space of the DSL, and a component to search in such a space. Next, we explain these components. 

\subsection{Learning Options}
\label{sec:options}

An option is a program encoding a policy that the agent can invoke in specific states. Once invoked, the program tells the agent what to do for a number of steps. Once completed, the option ``returns'' the control to the agent.
An option $\omega$ is defined with a tuple $(I_\omega, \pi_\omega, T_\omega)$, where $I_\omega$ 
is the set of states in which the option can be initiated; $\pi_\omega$ is the policy the agent follows once $\omega$ starts; $T_\omega$ is a function that returns the probability in which $\omega$ terminates at a given state. \name\ learns programs, written in a given DSL, encoding options. The programs receive a state of the MDP and return the action the agent should take at that state, thus encoding $\pi_{\omega}$. 

We assume that the set $I_{\omega}$ is the set $S$ of all states of the MDP, which means that the program can be invoked in any state. However, note that the program may not return any actions in a given state $s$, which is equivalent to $s$ not being in $I_{\omega}$. For example, ``if $b_1$ then $c_1$'' returns the action given in $c_1$ only if condition $b_1$ is satisfied in the state in which the option was queried; it returns no action for states that do not satisfy $b_1$. The option termination criterion $T_\omega$ is also determined by the program; the option terminates when the program terminates. This termination criterion means that, depending on the DSL, options have an internal state, representing the line in the program in which the execution will continue the next time the agent interacts with the environment. For example, if the option ``$c_1$ $c_2$'' is invoked for state $s_t$ and $c_1$ returns an action, then the agent's action in $s_{t+1}$ is determined by $c_2$. 

Programmatic options are harnessed from a foundation model as follows. We provide a natural language description of the MDP and the Backus-Naur form of the DSL to the model. The model then provides a set of $m$ programs written in the DSL encoding policies for the MDP. While it is only rarely that the model generates policies that maximize the expected return of the MDP, we hypothesize that these programmatic policies can be broken up into sub-programs that can encode options. Each program $p$ is broken into one sub-program for each sub-tree rooted at a non-terminal symbol in the AST of $p$. For example, for the program ``if $b_1$ then $c_1$ $c_2$'' we obtain the sub-programs ``if $b_1$ then $c_1$ $c_2$'', ``$b_1$'', ``$c_1$'', ``$c_2$'', and ``$c_1$ $c_2$''. These sub-programs form a set of options $O$, which \name\ uses to approximate the semantic space of the DSL. Note that this set of options is generated zero-shot, before the agent starts interacting with the environment.

\subsection{Inducing Semantic Spaces with Options}
\label{sec:semantic}

Methods for searching for programmatic policies traditionally search in the space of programs defined by the context-free grammar of the DSL~\cite{VermaMSKC18,carvalho2024reclaiming}. We refer to this type of space as the syntax space, since it is based on the syntax of the language. 

\begin{definition}[Syntax Space]
The syntax space of a DSL $D$ is defined by $(D, \mathcal{N}^x_k, \mathcal{I}, \mathcal{E})$. With $\llbracket D \rrbracket$ defining the set of candidate programs, or solutions, $\mathcal{N}^x_k$ ($x$ is for ``syntax'') is the syntax neighborhood function that receives a candidate and returns $k$ candidates from $\llbracket D \rrbracket$. $\mathcal{I}$ is the distribution of initial candidates. Finally, $\mathcal{E}$ is the evaluation function, which receives a candidate in $\llbracket D \rrbracket$ and returns a value in $\mathbb{R}$.
\label{definition:syntax}
\end{definition}

We define the distribution of initial candidates $\mathcal{I}$ through a procedure that starts with a string that is the initial symbol of the grammar and iteratively, and uniformly at random, samples a production rule to replace a non-terminal symbol in the string. In the example of Figure~\ref{fig:dsl}, we replace the initial symbol $\rho$ with \texttt{if($h$) then $a$} with probability $1.0$, since this is the only rule available; then, $h$ is replaced with \texttt{frontIsClear} or \texttt{markersPresent} with probability $0.5$ each. This iterative process stops once the string contains only terminal symbols. 

The syntax neighborhood function $\mathcal{N}^x_k$ defines the structure of the search space, as it determines the set of candidate solutions that the search procedure can evaluate from a given candidate $n$. Given a candidate $n$, $\mathcal{N}^x_k(n)$ returns a set of $k$ neighbors of $n$. These neighbors are generated by selecting, uniformly at random, a node that represents a non-terminal symbol in the AST of $n$, and then replacing the sub-tree rooted at the selected node by another randomly generated sub-tree. This process is repeated $k$ times, to generate $k$ possibly different neighbors of $n$. Finally, $\mathcal{E}$ is an approximation of the value function of the policy encoded in $n$ from a set of initial states $s_0$, $\hat{V}^n(s_0)$; we obtain $\hat{V}^n(s_0)$ by averaging the returns after rolling $n$ out from states $s_0$.

\citeauthor{semantic_spaces}~\shortcite{semantic_spaces} showed that searching in the syntax space can be inefficient because often the neighbors $n'$ of a candidate $n$ encode policies that are semantically identical to $n$; the programs differ in terms of syntax, but encode the same behavior. As a result, the search process wastes time evaluating the same agent behavior. They approximate the underlying semantic space of the DSL, where neighbor programs are syntactically similar but differ in terms of behavior. In their setting, the agent learns programs for a set of tasks, which are used to induce the semantic space, and the induced space is used in downstream tasks. \name\ overcomes the requirement to operate on a stream of problems by using a foundation model to learn the options in a zero-shot setting. 

\begin{definition}[Semantic Space]
The semantic space of a DSL $D$ is defined by $(D, \mathcal{N}^m_k, \mathcal{I}, \mathcal{E})$, where $\mathcal{I}$ and $\mathcal{E}$ are identical to the syntax space (Definition~\ref{definition:syntax}). The function $\mathcal{N}^m_k$ ($m$ is for ``semantics'') is a semantic neighborhood function that also receives a candidate and returns $k$ candidates from $\llbracket D \rrbracket$.
\end{definition}

We define the function $\mathcal{N}^m_k$ with a set of options $\Omega$, where each option $\omega$ in $\Omega$ represents a different agent behavior. A neighbor of candidate $n$ is obtained by selecting, uniformly at random, a node $c$ in the AST of $n$ that represents a non-terminal symbol. Then, we replace the sub-tree rooted at $c$ with the AST of an option $\omega$ in $\Omega$. The option $\omega$ is selected, uniformly at random, among those in $\Omega$ whose AST root represents the same non-terminal symbol $c$ represents. By matching the non-terminal symbols when selecting $\omega$, we match $\omega$ with the type of the sub-tree that is removed from $n$. Similarly to $\mathcal{N}^x_k$, $\mathcal{N}^m_k$ generates $k$ possibly different neighbors by repeating this process $k$ times. We require the options in $\Omega$ to encode different agent behaviors to increase the chance of sampling neighbors with different behaviors. 

We filter the set of options $O$ harnessed from the foundation model into a set $\Omega$ of behaviorally different options. First, we repeatedly sample an option $o \in O$, roll it out once in the environment (by playing itself in MicroRTS) from an initial state $s_0$ sampled from $\mu$, and collect all states observed in which the agent gets to act into a vector of states $\mathcal{S}$. We do this until we have at least 300 and at most 700 states in $\mathcal{S}$.\footnote{We chose these values so that we could approximate the behavior of the options well while being computationally reasonable.}  
Second, we test the options' behaviors by calling every option $o \in O$ once for each state $s \in \mathcal{S}$. We then collect the actions $o$ returns into a vector called an \emph{action signature} $A_o$ for the option $o$. The $i$-th entry of $A_o$ corresponds to the action $o$ chooses for the $i$-th state in $\mathcal{S}$. We use these action signatures to characterize the option's behavior. Finally, in order to form a set $\Omega$ of behaviorally different options, we keep only one option from $O$, arbitrarily selected, for each observed $A_o$. Note that this assumes discrete action spaces. Future research will investigate different ways of measuring behavior in continuous action spaces. 
Also, filtering $O$ into $\Omega$ uses less than 1\% of the computation in our experiments. Programs that cannot be rolled out (e.g., Boolean expressions cannot issue actions) are not included in the set of options. 

\subsection{Searching in Semantic Space}
\label{sec:searching}

\name\ uses stochastic hill-climbing (SHC) to search in the semantic space of a given DSL $D$ for a policy that maximizes the agent's return. SHC starts its search by sampling a candidate program $n$ from $\mathcal{I}$. In every iteration, SHC evaluates all $k$ neighbors of $n$ in terms of their $\mathcal{E}$-value. The search then moves on to the best neighbor of $n$ in terms of $\mathcal{E}$, and this process is repeated from there. SHC stops if none of the neighbors has an $\mathcal{E}$-value that is better than the current candidate, that is, it reaches a local optimum. SHC uses a restarting strategy: once SHC reaches a local optimum, if SHC has not yet exhausted its search budget, it restarts from another initial candidate sampled from $\mathcal{I}$. SHC returns the best solution, denoted $n^*$, encountered in all restarts of the search.  

\name\ does not search solely in the semantic space, but mixes both syntax and semantic spaces in the search. This is because the set of options might cover only a part of the space of programs the DSL induces. To guarantee that \name\ can access all programs in $\llbracket D \rrbracket$, with probability $\epsilon$, SHC uses the syntax neighborhood function in the search, and with probability $1 - \epsilon$, it uses the semantic one. We use $\epsilon = 0.4$ in our experiments. We chose this value because it performed better in preliminary experiments.

\section{Empirical Evaluation}

Although foundation models are unlikely to generate programs that encode strong policies, we hypothesize that the programs they generate can be broken into smaller programs encoding helpful options. We evaluated our hypothesis by measuring the sampling efficiency of algorithms searching in the semantic spaces induced by them. We evaluated \name\ on MicroRTS~\cite{ontanon2017combinatorial} and Karel the Robot~\cite{karel1994}.

\paragraph{MicroRTS}  We use the following maps from the MicroRTS repository,\footnote{https://github.com/Farama-Foundation/MicroRTS/} with the map size in brackets: NoWhereToRun ($9 \times 8$), 
basesWorkers ($24 \times 24$), 
and BWDistantResources ($32 \times 32$), and BloodBath ($64 \times 64$). We use these maps because they differ in size and structure. Since MicroRTS is a multi-agent problem, we use 2L, a self-play algorithm, to learn programmatic policies~\cite{locallearner}. In the context of 2L, \name\ is required to solve an MDP in every iteration of self-play (see Appendix~\ref{sec:local_learner}). We use a new version of the MicroLanguage as the DSL~\cite{marino2021programmatic}. The language offers specialized functions and an action-prioritization scheme through for-loops, where nested for-loops allow for higher priority of actions. We provide a detailed explanation of the MicroLanguage, as well as images of the maps used, in Appendices~\ref{sec:microlanguage} and \ref{sec:mrts_maps}, respectively.

\paragraph{Karel} 
We use the following Karel problems, from previous works~\cite{leaps,hprl}: StairClimber, FourCorners, TopOff, Maze, CleanHouse, Harvester, DoorKey, OneStroke, Seeder, and Snake. 
The problems are described in Appendix~\ref{sec:karel_problems}. We use the more difficult version of the environment known as ``crashable''~\cite{carvalho2024reclaiming}, where an episode terminates with a negative reward if Karel bumps into a wall. We use the same DSL used in previous work~\cite{leaps}, which we describe in Appendix~\ref{sec:karel_dsl}.  

\paragraph{Baselines} The current state-of-the-art methods for both MicroRTS and Karel use programmatic representations of policies, where the policies are written in the DSLs we use in our experiments~\cite{locallearner,leaps}. Therefore, we focus on methods that use programmatic representations as baselines. However, we provide comparisons of \name\ with deep reinforcement learning baselines in Appendices~\ref{sec:microrts_drl} (MicroRTS) and \ref{sec:karel_drl} (Karel). For both MicroRTS and Karel we use SHC searching in the syntax space as a baseline, as it represents state-of-the-art performance in both domains in our non-transfer setting (\textbf{SHC}). We also use two variants of \name\ where the options are learned without the help of a foundation model. These variants can be seen as implementations of the Library-Induced Semantic Spaces (LISS)~\cite{semantic_spaces} for non-transfer settings. In the first variant, LISS learns the options as it learns to solve the problem. In terms of the scheme shown in Figure~\ref{fig:innate_coder}, we skip the ``Learning Options'' step and build the set of options from the programs returned in every complete search of SHC. When we reach the box ``Restart?'', we use the sub-programs of the best program encountered in that search to augment the set of options. We call this baseline \textbf{LISS-o}, where ``o'' stands for ``online''. In the second variant, we sample programs from $\mathcal{I}$ and use their sub-programs as options. We call this baseline \textbf{LISS-r}, where ``r'' stands for ``random''. We also use the best program the foundation model generated from all programs used to create the set of options as a baseline called \textbf{FM}. LISS-o and LISS-r allow us to evaluate the effectiveness of learning options from a foundation model, while FM allows us to evaluate the foundation model as an alternative to solve the problem directly. We also use the Cross Entropy Method (\textbf{CEM}), which outperformed DRL algorithms in Karel~\cite{leaps}.

\paragraph{Foundation Models} 
We use OpenAI's API for GPT 4o, whose training cut-off date is October 2023. We also perform tests, for MicroRTS, using the LLama 3.1 model with 405 billion parameters, whose training cut-off is December 2021. We used the GPT model in both the MicroRTS and Karel experiments, while the Llama model was used in the MicroRTS experiments. There were no MicroRTS programs available online prior to the Llama cut-off date, so the Llama evaluations on MicroRTS did not suffer from data leakage. 
The GPT model might have trained on the MicroRTS and Karel programs that were available online prior to its training cut-off date. We attempt to measure how much a possible data leakage can influence our results by using the FM baseline. If the model can simply retrieve the solutions seen in training, one would expect this baseline to perform well. 

\paragraph{Other Specifications} All experiments were run on 2.6 GHz CPUs with 12 GB of RAM. We use $k = 1,000$ in the neighborhood function. In MicroRTS, SHC is run with a restarting time limit of $2,000$ seconds for each self-play iteration. In Karel, since we are solving a single MDP, SHC restarts as many times as possible within the computational budget. For MicroRTS, we query the foundation models $120$ times to generate the same number of programs; for Karel, we use $100$ programs. We use the same number of programs as the LISS-r baseline. We perform $30$ independent runs (seeds) of each system, including the generation of the programs by the model. 

\paragraph{\modtext{Metrics of Performance}} For MicroRTS, performance is measured in terms of winning rate: we sum the number of victories and half the number of draws and divide this sum by the total number of matches played~\cite{ontanon2017combinatorial}. For Karel, performance is measured in terms of episodic return~\cite{leaps}. We use prompts where we briefly describe each problem and provide a formal description of the DSL used. The prompts used in our experiments are in Appendices~\ref{sec:mrts_prompts} (MicroRTS) and \ref{sec:karel_prompts} (Karel). Both MicroRTS and Karel are deterministic, so the value of $\mathcal{E}$ for policies can be computed with a single roll-out. We report average performance and 95\% confidence intervals.

\paragraph{\modtext{Efficiency Experiment}} We verify the sampling efficiency of \name, LISS-o, LISS-r, and SHC. Similarly to previous work, we present learning curves, where for MicroRTS, we plot the winning rate by the number of games played (Figure~\ref{fig:microrts_results}), and for Karel, we plot episodic return by the number of episodes (Figure~\ref{fig:karel_results}). For MicroRTS, the winning rate is computed for a system by having the policy the system generated, after a given number of games played, play against the policies each of the other systems generated after the maximum number of games played (rightmost point of each plot). 

\paragraph{\modtext{Competition Experiment}}
We evaluate \name\ against COAC, Mayari, and RAISocketAI, the winners of the previous three MicroRTS competitions. We randomly select 9 from the 30 programs generated in the ``Efficiency Experiment'' and evaluate them against the competition winners. We report the average results of the 9 programs against each opponent in the four maps we use.

\paragraph{\modtext{Size and Information Experiments}} We also evaluate the effect of the size of the set of options on the sample efficiency of \name. We evaluated sets $\Omega$ with 300, 600, 1400, 5000, 7000 and 30000 options on the LetMeOut map (16 $\times$ 8); all options were generated with the Llama 3.1 model. 
In Appendix~\ref{sec:more_information}, we evaluate \name\ using prompts with more or less information and in Appendix~\ref{sec:model_size} with GPT 3.5, to verify if performance decreases if using a smaller model.

\subsection{Learning Curve Results}

\begin{figure*}[t]
    \centering
\includegraphics[width=0.95\textwidth]{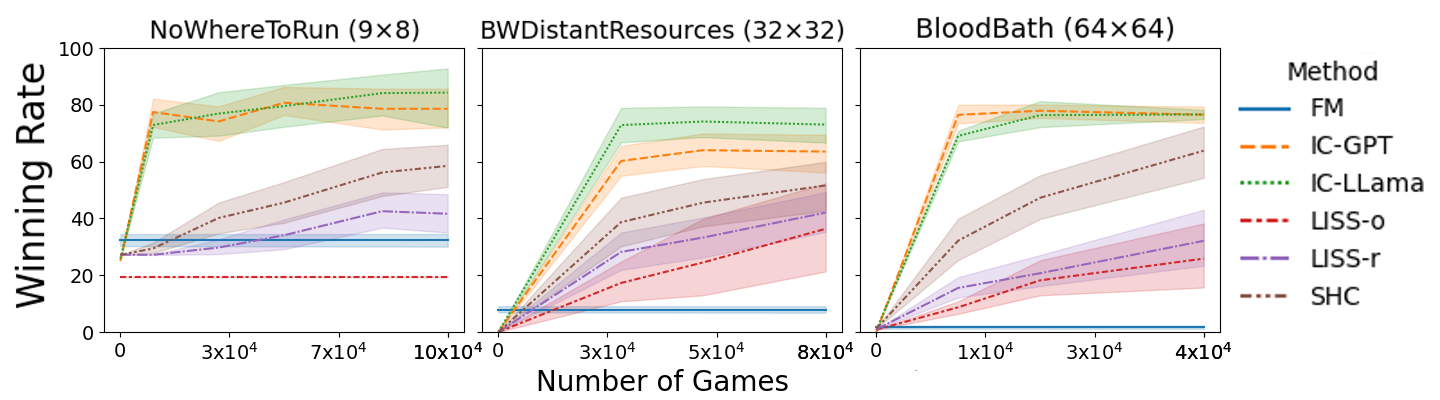}
    \caption{Winning rate (maximum is $100$) per number of games played. The winning rate of the policies each system generates for a given number of games played is computed considering as opponents the policies all systems generate at the end of the learning process. The plots show the average winning rate of 30 independent runs (seeds) and the 95\% confidence interval.}
    \label{fig:microrts_results}
\end{figure*}

\begin{figure*}[t]
    \centering
\includegraphics[width=0.90\textwidth]{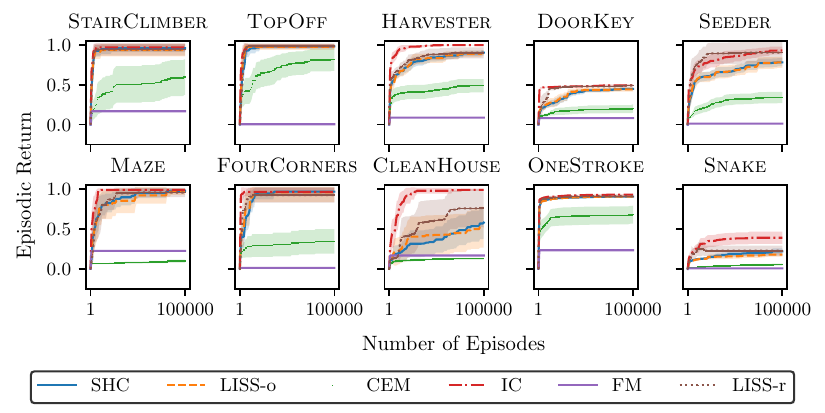}
    \caption{Average episodic return (maximum is $1.0$ for all tasks) per number of episodes. The plots show average episodic return of 30 independent runs (seeds) and the 95\% confidence interval.}
    \label{fig:karel_results}
\end{figure*}

Figures~\ref{fig:microrts_results} and \ref{fig:karel_results} show the learning curves for MicroRTS and Karel, respectively, where \name\ is denoted as IC-GPT or IC-Llama, depending on the model it uses to learn the options. \name\ is often much more sample-efficient than all baselines and, in many cases, by a large margin. We did not observe significant differences between IC-GPT and IC-Llama. LISS-o and LISS-r perform worse than \name\ and SHC in MicroRTS. However, LISS-o was competitive with SHC in Karel and LISS-r could outperform SHC (DoorKey and Seeder). The LISS-o and LISS-r results on MicroRTS suggest that the semantic space can be less conducive to search than the syntax space, depending on the quality of the options used to induce it. LISS-r performs better in Karel than in MicroRTS, probably because it uses a distribution $\mathcal{I}$ that uses a handcrafted probability distribution over the production rules of the language~\cite{leaps}. The resulting grammar allows for the generation of helpful options. 

FM performs poorly in all experiments.  
The results of FM and \name\ support our hypothesis that \name\ can extract helpful options from foundation models even if the programs the model generates do not encode strong policies. In MicroRTS, some of the options allowed the agent to allocate units to collect resources and train other units. Other systems had to learn such skills from scratch, while \name's agent had them ``innately available''. The FM results also suggest that data contamination was not an issue, as the model performed poorly on all tasks. 

\begin{table}[]
\centering
\footnotesize
\begin{tabular}{lrrrr}
\toprule
 \name\   & COAC    & RAI AI & Mayari  & Average    \\
  \midrule
GPT-4o      & 53.75 & 36.25     & 71.25 & 53.75 \\
Llama 3.1       & 43.79 & 70.00     & 58.17 & 57.32 \\
Llama 3.1 + GPT-4o & 70.14 & 72.92     & 46.39 & 63.15 \\
\bottomrule
\end{tabular}
\caption{Winning rate of \name\ against winners of previous competitions, averaged across all 4 maps used in our experiments.}
\label{tab:competition_microrts}
\end{table}

\subsection{Competition results}

Table~\ref{tab:competition_microrts} shows the results of \name\ against the winners of previous MicroRTS competitions, averaging its winning rate across 4 maps. COAC and Mayari are human-written policies, and RAISocketAI is a DRL agent~\cite{goodfriend2024competition}. RAISocketAI was trained with a larger computational budget than \name,  
thus giving it an advantage. We evaluated GPT-4o and Llama 3.1 while generating 120 programs. We also evaluate \name\ with the union of the programs generated by both GPT-4o and Llama (Llama 3.1 + GPT-4o in the table). The combination of GPT-4o and Llama 3.1 resulted in the best winning rate. 

The combination of programs written by Llama 3.1 and GPT-4o does not lead to ``monotonic improvements'', as evidenced by the drop in performance against Mayari. This happens because none of the competition winners is constrained by the DSL we use in our experiments. As a result, the optimization done in self-play might not be specific for the opponents evaluated in Table~\ref{tab:competition_microrts}, but to policies written in the DSL and encountered during the self-play process.

\subsection{Evaluating Number of Options}

\begin{figure}
    \centering
\includegraphics[width=0.85\columnwidth]{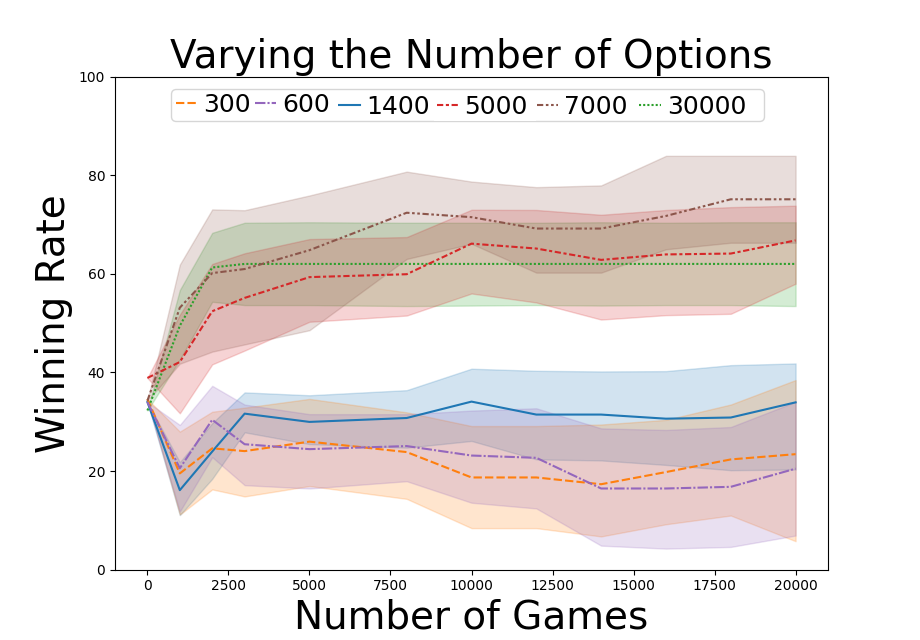}
    \caption{Average winning rate of \name\ policies for different sizes of the option set over 10 independent runs (seeds) of each version. We also present the 95\% confidence intervals.}
    \label{fig:varying_number_options}
\end{figure}

Figure~\ref{fig:varying_number_options} presents the learning curves for different versions \name, where we vary the size of the set of options $\Omega$. The three lines with the highest winning rate are for option sets of sizes 5000, 7000, and 30000. The versions of \name\ with option sets of sizes 300, 600, and 1400 perform worse. These results demonstrate that \name\ can benefit from thousands of options. This is possible due to \name's way of using options through the induction of the language's underlying semantic space. 

These results also show that \name's sample efficiency plateaus at 5000 options, since the use of 7000 and 30000 options does not increase performance. Interestingly, performance does not degrade either as we increase the set size. We conjecture that this occurs because many of the options will encode different and yet similar behaviors that do not affect the agent's winning rate. 
Although the set of distinct behaviors encoded in the set options grows with larger sets, the relative number of options with behaviors that affect the winning rate remains roughly the same. As a result, the function $\mathcal{N}^m_k$ that uses option sets of sizes 5000, 7000, or 30000 induces spaces that are similarly conducive to search.

\section{Related work}

\paragraph{Programmatic Policies} One of the key challenges in generating programmatic policies is that the search space is discontinuous and gradient-based optimization cannot be used. Some previous work relied on imitation learning to guide the search for policies~\cite{VermaMSKC18,verma2019imitation,viper}. 
The issue of this imitation learning approach is known as \emph{representation gap}~\cite{QiuZ22,medeiros2022can}, where the space of programmatic policies does not include the oracle policy that the system tries to imitate. As a result, the oracle might guide the search to unpromising parts of the space. Previous work tried to learn latent spaces of programming languages that are conducive to search~\cite{leaps,hprl}, which was shown to be outperformed by the syntax space with SHC~\cite{carvalho2024reclaiming}. Semantic spaces were shown to be more conducive to search than syntax spaces, but required a sequence of tasks, where the agent learns the space in one task and reuses it in others~\cite{semantic_spaces}. Our work does not require an oracle nor a sequence of tasks to learn the semantic space, which is learned in a zero-shot setting. 

\paragraph{Options} Options were shown to improve the sampling efficiency of learning agents through faster credit assignment~\cite{pmlr-v32-mann14}, better exploration~\cite{BARANES201349,bellemare2020autonomous}, and transfer of knowledge across tasks~\cite{konidaris2007building,alikhasi2024unveiling}. However, previous methods for learning options require the user to design them before learning starts~\cite{options_sutton} or to provide considerable information as input to the process, such as the option duration~\cite{frans2017meta,tessler2017deep} or the number of options learned~\cite{bacon2017option,igl2020multitask}. Other methods rely on the agent interaction with the current environment~\cite{achiam2018variational,machado2018eigenoption,jinnai2020exploration} or with other earlier environments, as in transfer learning approaches~\cite{konidaris2007building,alikhasi2024unveiling}. We present a novel way of learning options as they are not learned from the agent's experience nor designed by the user, but harnessed from foundation models. While we use options to define a search space, future work will explore their use as functions neural policies can call.

\section{Conclusions}

In this paper, we introduced \name, a system that equips learning agents with skills, in the form of programmatic options, before the agent starts to interact with the environment. This is achieved by extracting programmatic options from foundation models. We hypothesized that even if the model is unable to write programs encoding strong policies for a problem, sub-programs of the generated program could encode helpful options. We tested our hypothesis in MicroRTS and Karel, two domains in which programmatic policies represent the current state of the art. The policies \name\ generated outperformed, often by a large margin, a baseline that did not attempt to learn options; a baseline that learned the options while learning how to solve the problem; a baseline that learned the options from programs sampled directly from the domain-specific language; and the foundation model that attempted to generate programmatic policies directly. We also showed that some of the policies \name\ generated were competitive or outperformed the winners of previous MicroRTS competitions, including programmatic policies written by human programmers and a deep reinforcement learning agent that used a larger computational budget than we allowed \name\ to use. These results place \name\ as the current state-of-the-art in both Karel and MicroRTS. Our experiments also showed that \name's scheme of using programmatic options to induce semantic spaces allows it to benefit from thousands of options, while most previous work can benefit only from dozens or at most hundreds of options~\cite{eysenbach2019diversity}.

\section*{Acknowledgments}

This research was supported by Canada's NSERC, and the CIFAR AI Chairs program, and Brazil's CAPES. The research was carried out using computational resources from the Digital Research Alliance of Canada and the UFV Cluster. We thank the anonymous reviewers for their feedback.

\bibliographystyle{named}
\bibliography{ijcai25}

\newpage
\appendix
\onecolumn 

\section{Additional Related Work}
\label{sec:additional_related_work}

\paragraph{Foundation Models as Policies} Foundation models have been used to perceive, plan, and act \cite{DBLP:conf/uist/ParkOCMLB23}, often decomposing long-horizon goals into subtasks \cite{DBLP:journals/corr/abs-2302-01560}, and/or integrating additional agent features such as memory \cite{DBLP:journals/corr/abs-2305-17144} and/or automatic learning curricula \cite{DBLP:journals/corr/abs-2305-16291}. By contrast, \name\ uses the model as a pre-processing step to generate programmatic options, which make it more accessible due to the limited number of model calls. Foundation models have also been used to learn reward functions that are later used to train agents~\cite{klissarov2024motif}. Similarly to our work, the model is used in a pre-processing step. However, in contrast with our work, it learns reward functions, while we learn programmatic options. Moreover, it needs a ``diverse'' set of states, which are generated by existing and proficient agents; \name\ learns in a zero-shot setting and only uses the generated options themselves to generate a set of states used to filter out the options encoding non-novel behaviors.

\paragraph{Foundation Models for Planning} Previous work has used foundation models for generating programmatic policies in the context of planning. For example, in generalized planning (GP), the goal is to synthesize programs that solve classical planning problems \cite{DBLP:journals/ker/CelorrioAJ19}; foundation models have shown promise for GP \cite{DBLP:journals/corr/abs-2305-11014}. Foundation models has also been used successfully in the context of code generation for decision making in robotics \cite{DBLP:conf/icra/LiangHXXHIFZ23,DBLP:conf/icra/SinghBMGXTFTG23}. In contrast to these works, we do not attempt to use the model's generated program as a policy, but we extract options from them and use these options to induce a search space. 

\paragraph{Foundation Models as Search Guidance} Foundation models have also been used to guide search algorithms. This includes methods for solving optimization problems \cite{DBLP:journals/corr/abs-2309-03409,DBLP:journals/corr/abs-2310-05204} and to guide Monte Carlo tree search~\cite{DBLP:journals/corr/abs-2305-14078}. Foundation models were also used in genetic operators~\cite{DBLP:journals/corr/abs-2206-08896,DBLP:journals/corr/abs-2310-19046,DBLP:journals/corr/abs-2302-12170,DBLP:journals/corr/abs-2302-14838}, including multi-objective \cite{DBLP:journals/corr/abs-2310-12541} and quality-diversity algorithms \cite{DBLP:journals/corr/abs-2306-01102}.  These works are resource-intensive  due to calling the model during the search~\cite{DBLP:journals/corr/abs-2310-12541}. This contrasts with our work, which uses the model a small number of times in a pre-processing step.

\section{Deep Reinforcement Learning Comparison}

\subsection{MicroRTS}
\label{sec:microrts_drl}

To compare \name\ with a Deep Reinforcement Learning algorithm, we used the Gym-MicroRTS \cite{huang2021gym}. We evaluated \name\ with PPO Gridnet self-play using an encode-decode model. We chose this model because is the closest one to ours, as they both learn through self play. The DRL agents proposed by \citeauthor{huang2021gym} were specifically designed and tested for a map of size 16 $\times$ 16. We used the Gym-MicroRTS with the same settings presented in the repository, changing only the budget and the UnitTable used in the experiments. We trained both algorithms with a budget of 300 million steps in the MicroRTS. 

\begin{figure}[ht]
    \centering
    \includegraphics[width=.48\textwidth]{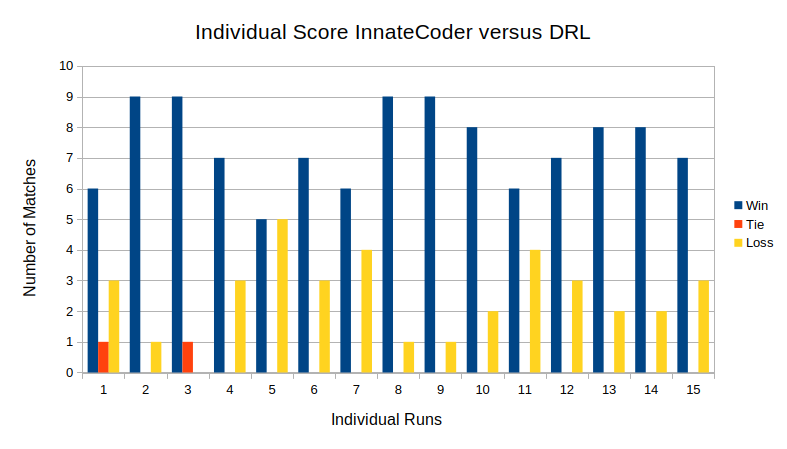}
    \caption{Individual results for \name\ against DRL in basesWorkers16x16A map. The columns are in order: win, tie, and loss, that \name\ got against DRL.}
    \label{fig:drlindividual}
\end{figure}

\begin{figure}[ht]
    \centering
    
    \includegraphics[width=.48\textwidth]{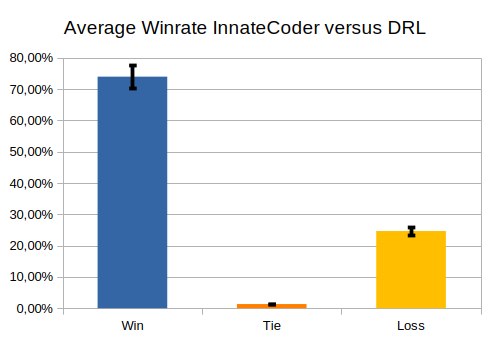}
    \caption{Average results of \name\ against DRL in basesWorkers16x16A map, with 95\% confidence interval.}
    \label{fig:drlavg}
\end{figure}

We trained 15 DRL models for the BasesWorkers map (16$\times$16), using a Xeon 2.90GHz, 64GB of memory, and a dedicated Nvidia A10. Also, we performed 15 individual runs for \name\. For each pair (DRL-\name) we ran 10 matches of the policies using the same evaluation used by \citeauthor{huang2021gym}. Each individual result is shown in Figure~\ref{fig:drlindividual}. Each run shows the number of losses, ties, and wins that \name achieved against DRL. The best score obtained by DRL is 5 losses and 5 wins presented in the individual run number 5 of graphs. In contrast to the policies \name\ generates, which can be used to play at any of the two locations of the map, the DRL-PPO agent is trained for a fixed position. This provides an advantage to DRL-PPO, as \name\ does not specialize in a given location of the map. Figure \ref{fig:drlavg} shows the average results, where the whiskers show the 95\% confidence interval. \name\ wins more than 70\% of the matches with the PPO agent. Moreover, while the DRL agent needs around 3 days to train for 300 million steps, \name\ can perform the same number of training in less than 36 hours of computation using a single CPU. 

\subsection{Karel}
\label{sec:karel_drl}

For Karel the Robot, we compared \name\ with Hierarchical Programmatic Reinforcement Learning (HPRL)~\cite{hprl}, which uses neural and programmatic representations of policies. We evaluate the version of the system that uses PPO (HPRL-PPO), one of the best variants of HPRL. Table \ref{tab:karel_hprl_ppo} shows the comparison between \name, HPRL-PPO, SHC~\cite{carvalho2024reclaiming}, and CEM~\cite{leaps}. These tests were performed with a budget of $10^5$ episodes. \name\ (IC in the table) obtained the highest average return in all tasks. 

\begin{table}[ht]
\footnotesize
\setlength{\tabcolsep}{1.8pt}
\centering
\begin{tabular}{lcccc}
\toprule
Task               & HPRL-PPO    & SHC         & CEM         & IC         \\
\midrule
StairClimberSpar. & \textbf{1.00}$\pm$0.00   & \textbf{1.00}$\pm$0.00   & 0.60$\pm$0.44 & \textbf{1.00}$\pm$0.00   \\
MazeSparse         & \textbf{1.00}$\pm$0.00   & \textbf{1.00}$\pm$0.00   & 0.09$\pm$0.03 & \textbf{1.00}$\pm$0.00   \\
TopOff             & \textbf{1.00}$\pm$0.00   & \textbf{1.00}$\pm$0.00   & 0.81$\pm$0.29 & \textbf{1.00}$\pm$0.00   \\
FourCorners        & \textbf{1.00}$\pm$0.00   & \textbf{1.00}$\pm$0.00   & 0.33$\pm$0.29 & \textbf{1.00}$\pm$0.00   \\
Harvester          & 0.92$\pm$0.13   & 0.90$\pm$0.07   & 0.48$\pm$0.17 & \textbf{1.00}$\pm$0.00   \\
CleanHouse         & 0.82$\pm$0.21   & 0.59$\pm$0.41   & 0.12$\pm$0.02 & \textbf{1.00}$\pm$0.00   \\
DoorKey            & 0.38$\pm$0.09   & 0.44$\pm$0.04   & 0.20$\pm$0.11 & \textbf{0.49}$\pm$0.03 \\
OneStroke          & 0.78$\pm$0.11   & 0.90$\pm$0.01   & 0.68$\pm$0.23 & \textbf{0.93}$\pm$0.01 \\
Seeder             & 0.53$\pm$0.17   & 0.77$\pm$0.10   & 0.33$\pm$0.14 & \textbf{0.93}$\pm$0.09 \\
Snake              & 0.28$\pm$0.18   & 0.21$\pm$0.08   & 0.05$\pm$0.02 & \textbf{0.39}$\pm$0.15\\
\bottomrule
\end{tabular}
\caption{Mean and standard error of final episodic return of \name\, HPRL-PPO, SHC, and CEM in Karel problems.}
\label{tab:karel_hprl_ppo}
\end{table}

\section{Can \name\ Improve with more Information?}
\label{sec:more_information}

\begin{figure}[t]
    \centering
\includegraphics[width=1.0\columnwidth]{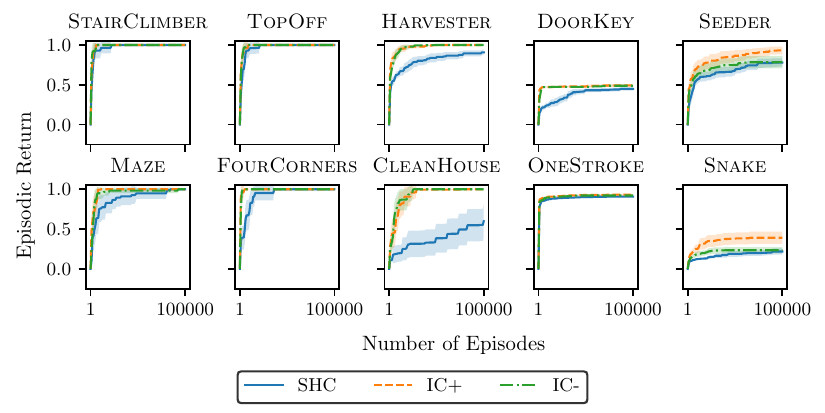}
    \caption{Evaluating \name\ with more (IC+) and less (IC-) information provided in the prompt used to harness programmatic options from the foundation model. We used GPT 4o in this experiment. Average episodic return (maximum is $1.0$ for all tasks) per number of episodes. The plots show average episodic return of 30 independent runs (seeds) and the 95\% confidence interval.}
    \label{fig:karel_results_more_less}
\end{figure}

We evaluated whether \name's sample efficiency can scale with the amount of information we provide in the prompt used to generate the programmatic options. Figure~\ref{fig:karel_results_more_less} shows the results of two versions of \name: one that uses prompts with more information (IC+) and one that uses prompts with less information (IC-). The prompts are given in Section~\ref{sub:more_info} (more information) and Section~\ref{sub:less_info} (less information). The key difference between IC+ and IC- is that, in the former, we explain in the prompt how the agent can maximize its return. For example, in FourCorners we wrote ``agent has to place one
marker in each of the four corner cells of the grid''. In contrast, in IC- we wrote ``the robot will receive different reward values depending on its interactions inside the grid''. Providing more information was never worse, and it was significantly better in two cases: Seeder and Snake. The ability to improve with more information is important because it allows the user of \name\ to achieve stronger results by crafting prompts that encode domain knowledge. 

\section{Can \name\ Improve with Model Size?}
\label{sec:model_size}

\begin{figure}[h]
    \centering
\includegraphics[width=1.0\columnwidth]{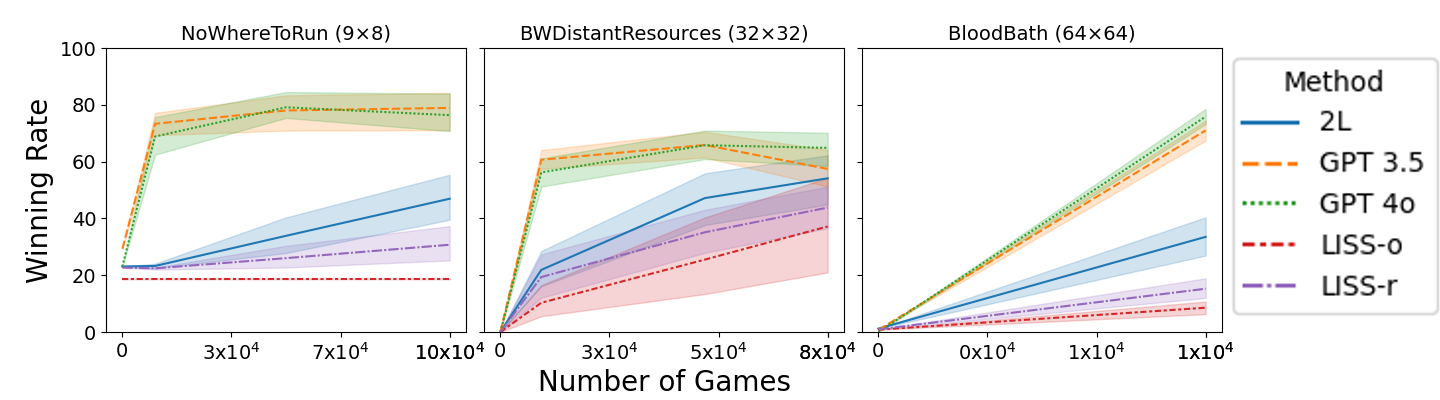}
    \caption{ The plots show the average winning rate of 30 independent runs (seeds) and the 95\% confidence interval.}
    \label{fig:gpts}
\end{figure}

We evaluated GPT 3.5-turbo and GPT 4o on the MicroRTS tasks. Figure~\ref{fig:gpts} shows the results, where we report the average winning rate and the 95\% confidence intervals of 30 independent runs (seeds). Interestingly, we do not notice a significant change in winning rate as we move from the larger GPT 4o to the smaller GPT 3.5. 

\section{\modtext{Samples of Programs}}

\modtext{Figure~\ref{fig:samples_programs} shows an example of a programmatic policy \name\ generated for the BloodBath 64$\times$64 map (left), and a program Llama 3.1 generated for the same map (right). The policy \name\ generated defeats the last three winners of the MicroRTS competition: COAC, Mayari, and RAISocketAI. This policy presents non-trivial features. For example, lines 7-11 will train Worker and Ranged units only if the player is not engaged in combat. This means that this policy focuses on economy and on Ranged units in the early stages of the game. Later in the game, the agent will save its resources to train Light units (line 21). Light units can be trained and move more quickly than Ranged units. While a Light unit is trained in 80 time steps of the game, a Ranged unit requires 100 time steps to be trained; also, a Light unit can move one cell every 8 time steps, while Ranged units move one cell every 10 time steps. Light units allow for a faster return of the resources invested.}

\modtext{The program the foundation model generates represents a weak policy (program shown on the right of Figure~\ref{fig:samples_programs}). However, even weak policies can contain pieces of code---options---that can be helpful while searching for strong policies to play the game. For example, lines 2 and 3 offer a prioritization scheme for investing resources to train Worker units because it will iterate over all units until it finds the Base, which will be used to train up to 5 workers. Lines 2 and 3 are often found in strong policies. For example, the program shown on the left of Figure~\ref{fig:samples_programs} shows a similar structure in lines 1 and 10.}

\begin{figure}[t]
    \centering
    \begin{minipage}[t]{0.48\textwidth}
    \begin{lstlisting}[language=Python]
# InnateCoder's policy for the 64x64 map
for (Unit u)
    u.attackIfInRange()
    for (Unit u)
        u.attackIfInRange()
        u.train(Heavy, EnemyDir, 50)
    u.harvest(10)
    if (u.OpponentHasUnitInPlayerRange()) 
        pass
    else
        u.train(Worker, EnemyDir, 3)
        u.train(Ranged, EnemyDir, 15)
    u.attack(Strongest)
    for (Unit u) 
        u.harvest(5)
    for (Unit u) 
        u.build(Barracks, EnemyDir, 8)
    u.moveToUnit(Enemy, Strongest)
    if (u.HasUnitInOpponentRange())
        for (Unit u) 
            u.moveToUnit(Enemy, Farthest)
            u.train(Light, EnemyDir, 6)
    u.train(Worker, Down, 6)
    \end{lstlisting}
    \end{minipage}
    \begin{minipage}[t]{0.48\textwidth}
    \begin{lstlisting}[language=Python]
# Llama's policy for the 64x64 map
for(Unit u) 
  for(Unit u) 
    u.train(Worker, Down, 5)
  for(Unit u) 
    u.moveToUnit(Ally, Closest)
    u.harvest(5)
  for(Unit u) 
    u.build(Barracks, Up, 1)
    u.train(Ranged, EnemyDir, 10)
    u.attackIfInRange()
  for(Unit u) 
    u.moveToUnit(Enemy, Weakest)
    u.attack(Weakest)
    \end{lstlisting}
    \end{minipage}
    \caption{\modtext{Left: A programmatic policy \name\ generated for the largest 64$\times$64 map. This policy defeats the last three winners of the MicroRTS Competition: COAC, Mayari, and RAISocketAI. Right: one of the policies Llama 3.1 generated for the same map.}}
    \label{fig:samples_programs}
\end{figure}

\section{\modtext{Option Usage in Programmatic Policies}}
\label{sec:option_usage}

\begin{figure}[t]
    \centering
    \includegraphics[width=0.5\linewidth]{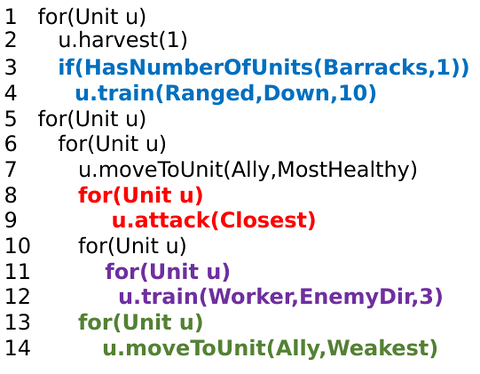}
    \caption{\modtext{Programmatic policy \name\ generated for the LetMeOut map (16 $\times$ 8). The colored lines represent options from one execution of \name\ that used 5000 options.}}
    \label{fig:option_usage}
\end{figure}

\modtext{Figure~\ref{fig:option_usage} shows an example of a program \name\ generated as the final policy (the one that the system produces as output) for the LetMeOut map (16 $\times$ 8), where four options are used. The colored lines represent the options. Lines 3 and 4 form one option (blue), while lines 8 and 9 (red), 11 and 12 (purple), and 13 and 14 (green) form the other three options. 
In this representative example, 8 of 14 lines are options, which represents 57\% of the lines.  
We also analyzed 10 independent runs of \name\ in the 9$\times$8 map with an initial option set extracted from 120 policies, which were generated with the Llama 3.1 model. We found that on average, 63\% of the options in the initial set are used in a best response during the self-play training process. Recall that a best response is the solution \name\ finds to a given MDP within the self-play algorithm. The maximum percentage of options used in the 10  runs was 73\% and the minimum was 46\%. Even if an option is not used in the policy \name\ outputs, the option can still have played an important role in allowing \name\ arrive at the output it produced. This is because an option could have been part of one of the best responses encountered during self play, and these best responses provide the signal needed to guide the search toward stronger policies for playing the game~\cite{locallearner}. 
}

\modtext{The policy shown in Figure~\ref{fig:option_usage} also provides an explanation for the difference in performance between the versions of \name\ that are initialized with a pool of 1400 or fewer options and the versions that are initialized with a pool of 5000 or more options (see Figure~\ref{fig:varying_number_options}). The options highlighted in Figure~\ref{fig:option_usage} are in the set of 5000 options of an \name\ run, but they are not in the set of 1400 options of another run of the algorithm. These options are clearly helpful because they appear in the output policy. Sampling 5000 options instead of 1400 increases the chances of adding some of these helpful options to the library, thus explaining the gap we see between \name\ with 1400 options or fewer and \name\ with 5000 options or more in Figure~\ref{fig:varying_number_options}.}

\section{\modtext{Varying the Value of $\epsilon$}}

\begin{figure}[t]
    \centering
    \includegraphics[width=0.7\linewidth]{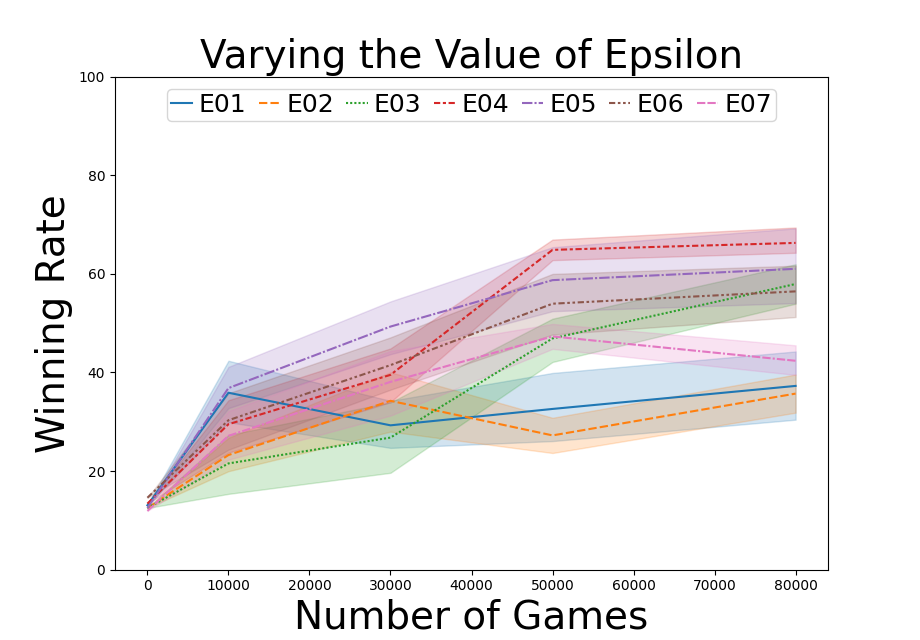}
    \caption{\modtext{\name\ with values of $\epsilon$ in $\{0.1, 0.2, 0.3, 0.4, 0.5, 0.6, 0.7\}$ in the 9$\times$8 MicroRTS map. Here, each \name\ line is evaluated as in the ``Efficiency Experiment'' (Figure~\ref{fig:microrts_results}). Each line shows the average of 30 independent runs and the 95\% confidence intervals.}}
    \label{fig:varying_epsilon}
\end{figure}

\modtext{Figure~\ref{fig:varying_epsilon} shows the performance of \name\ as we vary the value of $\epsilon$. Recall that $\epsilon$ dictates how much of the search is done in the syntax space and how much of the search is done in the semantic space. Also, recall that smaller values of $\epsilon$ mean that the semantic space is used more often. There are two groups of lines at 80000 games played: $\epsilon$-values of $0.3$, $0.4$, $0.5$, and $0.6$ (top lines) and $0.1$, $0.2$, and $0.7$ (bottom lines). These results suggest that \name\ is robust to the choice of $\epsilon$ value, as a wide range of values produce competitive results among themselves. The results also suggest that if the value of $\epsilon$ is too small, then the search does not sufficiently explore the syntax space to eventually find programs that were not originally in the library of options. If the value of $\epsilon$ is too large, then the search is not making use of the helpful options in the library as often as it could.} 

\section{Domain-Specific Languages (DSLs)}

In this section, we present the DSLs used in our experiments for both Karel the Robot and MicroRTS.

\subsection*{Karel}
\label{sec:karel_dsl}

The context-free grammar below presents the DSL for Karel the Robot. This DSL is the same as that used in the previous work~\cite{leaps,hprl}.

\begin{align}
    \text{Program }\rho :=~& \texttt{DEF run m( } s \texttt{ m)} \\
    \text{Statement }s := &\texttt{WHILE c( }b\texttt{ c) w( }s\texttt{ w) } |\\ 
    & \texttt{IF c( }b\texttt{ c) i( }s\texttt{ i) } | \\
    & \texttt{IFELSE c( }b\texttt{ c) i( }s\texttt{ i) ELSE e( }s\texttt{ e) } | \\
    & \texttt{REPEAT R=}n \texttt{ r( }s\texttt{ r) } | \\
    & s;s\texttt{ }|\texttt{ }a \\
    \text{Condition }b :=~& h\texttt{ }| \texttt{ not( }h\texttt{ )} \\
    \text{Number } n :=~& 1~|~2~|~3~|~...~|~\texttt{infinity} \\
    \text{Perception }h :=~& \texttt{frontIsClear }|\texttt{ leftIsClear }|\texttt{ rightIsClear }| \\
    & \texttt{markersPresent }|\texttt{ noMarkersPresent } \\
    \text{Action }a :=~& \texttt{move }|\texttt{ turnLeft }|\texttt{ turnRight }| \texttt{ putMarker }|\texttt{ pickMarker }
\end{align}

\subsection*{MicroRTS}\label{sec:microlanguage}

The context-free grammar below presents the DSL for MicroRTS. This DSL is the same as that used in recent work~\cite{semantic_spaces}. Note that an early version of the Microlanguage appears in the work of \cite{marino2021programmatic}, which was published prior to the cut-off date of the ChatGPT model used in our experiments. However, this earlier version of the language is fundamentally different from the one used in recent work and in our experiments. For context, the interpreter we use in our experiments cannot run the programs written in the language used of \cite{marino2021programmatic}. To illustrate some differences, the instruction \texttt{moveToUnit(Light, Ally, strongest, u)} in the older Microlanguage has four parameters, while in ours it has only two. The older language has a larger collection of high-level functions, such as \texttt{HaveQtdUnitsbyType}, which our version of the language does not have. Considering these key differences, it is unlikely that the data from \cite{marino2021programmatic} have influenced our results more than the vast collection of programs written in various programming languages that are present in the corpus used to train these foundation models. 

\begin{align*}
        S &\to SS \mid \texttt{for(Unit u) S} \mid \texttt{if(B) then S} \\
        &\quad \mid \texttt{if(B) then S else S} \mid C \mid \lambda \\
        B &\to \texttt{hasNumberOfUnits}(T, N) \mid \texttt{opponentHasNumberOfUnits}(T, N) \\
        &\quad \mid \texttt{hasLessNumberOfUnits}(T, N) \mid \texttt{haveQtdUnitsAttacking}(N) \\
        &\quad \mid \texttt{hasUnitWithinDistanceFromOpponent}(N) \\ &\quad \mid \texttt{hasNumberOfWorkersHarvesting}(N) \\
        &\quad \mid \texttt{is\_Type}(T) \mid \texttt{isBuilder()} \\
        &\quad \mid \texttt{canAttack()} \mid \texttt{hasUnitThatKillsInOneAttack()} \\
        &\quad \mid \texttt{opponentHasUnitThatKillsUnitInOneAttack()} \\
        &\quad \mid \texttt{hasUnitInOpponentRange()} \\
        &\quad \mid \texttt{opponentHasUnitInPlayerRange()} \\
        &\quad \mid \texttt{canHarvest()}\\
        C &\to \texttt{build}(T, D, N) \mid \texttt{train}(T, D, N) \mid \texttt{moveToUnit}(T_p, O_p) \mid \texttt{attack}(O_p)  \\
        &\quad \mid \texttt{harvest}(N) \mid \texttt{attackIfInRange()} \mid \texttt{moveAway()}\\
        T &\to \texttt{Base} \mid \texttt{Barracks} \mid \texttt{Ranged} \mid \texttt{Heavy} \\
        &\quad \mid \texttt{Light} \mid \texttt{Worker} \\
        N &\to 0 \mid 1 \mid 2 \mid 3 \mid 4 \mid 5 \mid 6 \mid 7 \mid 8 \mid 9 \\
        &\quad \mid 10 \mid 15 \mid 20 \mid 25 \mid 50 \mid 100\\
        D &\to \texttt{EnemyDir} \mid \texttt{Up} \mid \texttt{Down} \mid \texttt{Right} \mid \texttt{Left}\\
        O_p &\to \texttt{Strongest} \mid \texttt{Weakest} \mid \texttt{Closest} \mid \texttt{Farthest} \\
        &\quad \mid \texttt{LessHealthy} \mid \texttt{MostHealthy} \mid \texttt{Random}\\
        T_p &\to \texttt{Ally} \mid \texttt{Enemy}\\
\end{align*}

\section{\textsc{Karel Problem Sets}} \label{sec:karel_problems}
The \textsc{Karel} problem sets \cite{leaps,hprl} are divided into two parts--- \textsc{Karel} and \textsc{Karel-Hard}. \textsc{Karel} consists of six different tasks, while \textsc{Karel-Hard} includes four additional tasks that are comparatively more difficult to solve. In this section, we describe the initial state and the return function of each task in both \textsc{Karel} and \textsc{Karel-Hard} problem sets.

\subsection{\textsc{Karel}}

\paragraph{StairClimber.}
In this task, the agent operates within a $12\times12$ grid containing stairs formed by walls. The goal for the agent is to reach a marker located above its position on the stairs. The initial positions of the agent and the marker are randomly initialized on the stairs. The agent receives an episodic return of $1$ if it successfully reaches the marker, and $0$ otherwise. Moving to a position outside the contour of the stairs results in a return of $-1$.

\paragraph{FourCorners.}
In this task, the goal for the agent is to place a marker in the four corners of a $12\times12$ grid. The initial position of the agent is randomly initialized near the wall. The return is calculated as the number of corners with one marker divided by four.

\paragraph{TopOff.}
In this task, the agent is always initialized on the bottom left of a $12\times12$ grid, and the markers are placed randomly on the bottom row of the grid. The goal for the agent to place markers on top of the markers present in that row. The return is calculated as the number of markers topped off divided by the number of markers present in the grid.

\paragraph{Maze.}
A maze, formed by walls, is randomly configured on a $12\times12$ grid, and a marker and an agent are randomly placed within it. The goal for the agent is to reach the marker, that yields an episodic return of $1$. Otherwise, the episodic return is $0$.

\paragraph{CleanHouse.}
Markers and walls are randomly placed in a $22\times14$ grid, referred to as the apartment. The position of the agent is also initialized randomly. Its goal is to pick all the markers inside the apartment. The return is calculated as the number of picked markers divided by the total number of markers present.

\paragraph{Harvester.}
In this task, the agent operates within an $8\times8$ grid filled with markers. The agent is placed randomly on the bottom row, with the goal of collecting all markers in the grid. The return is calculated as the number of collected markers divided by the total number of cells in the grid.

\subsection{\textsc{Karel-Hard}}

\paragraph{DoorKey.}
The environment, consisting of an $8\times8$ grid is divided into two chambers by a vertical bar with a door. The position of the agent is initialized randomly, and a marker is placed randomly in each chamber. The goal for the agent is to pick the marker in the left chamber, which will open the door in the vertical bar, and then pick the marker in the right chamber. Picking each marker results in a return of $0.5$.

\paragraph{OneStroke.}
In this task, the goal for the agent is to visit every cell of an $8\times8$ grid without repetition. Once a cell is visited, it transforms into a wall, and the episode terminates if the agent hits a wall. The return is calculated as the number of visited cells divided by the total number of cells in the grid.

\paragraph{Seeder.}
The environment is given by an $8\times8$ empty grid with an agent initialized at a random position. The goal for the agent is to place a marker in each cell of the grid. The return is calculated as the number of placed markers divided by the total number of cells.

\paragraph{Snake.}
In this task, an $8\times8$ grid is initialized with an agent and a marker at random positions. The agent behaves like the head of a snake, with its body growing after collecting a marker. Once a marker is collected, another marker is placed at a random position. This process continues until the agent collects $20$ markers. The goal for the agent is to collect the markers without hitting its own body. The return is calculated as the number of collected markers divided by $20$.

\section{\textsc{Karel Prompts}} \label{sec:karel_prompts}
In this section, we present the prompts used to generate a program that encodes a policy. These are the prompts we used multiple times to obtain a set of programmatic policies. This section is divided into three subsections. Section~\ref{sub:full_prompt} includes the complete prompt used to obtain a program from the model to solve the `Seeder' task. For each task, we provide two types of prompts: one with more information and another with less information. Section~\ref{sub:more_info} contains the details of the environment with more information for each task, while Section~\ref{sub:less_info} contains the prompts with less information for each task.

\subsection{\textsc{Complete Prompt for the Seeder Task}} \label{sub:full_prompt}

The prompt for obtaining a program that encodes a policy for the `Seeder' task is shown below. Note that the first paragraph of the prompt explains the environment. Only this paragraph is changed from one task to the next.

\begin{tcolorbox}[breakable, width=\textwidth, colback=white]
    Consider an 8x8 gridded 2D environment in Karel the Robot, where an agent has to place one marker in every cell of the grid. The grid is initially empty and the initial position of the agent is randomly assigned at the beginning of each episode.\\

    The following is the Context Free Grammar (CFG) for the Karel domain:
    \begin{align*}
        P &\to \text{run \{S\}}\\
        S &\to \text{WHILE (B) \{S\}} \GOR \text{S S} \GOR \text{A} \GOR \text{REPEAT R=N \{S\}} \GOR \text{IF (B) \{S\}} \GOR
        \text{IF (B) \{S\} ELSE \{S\}}\\
        B &\to \text{H} \GOR \text{not H}\\
        H &\to \text{frontIsClear} \GOR \text{leftIsClear} \GOR \text{rightIsClear} \GOR \text{markersPresent} \GOR \text{noMarkersPresent}\\
        A &\to \text{move} \GOR \text{turnLeft} \GOR \text{turnRight} \GOR \text{putMarker} \GOR \text{pickMarker}\\
        N &\to 1 \GOR 2 \GOR 3 \GOR ... \GOR \text{infinity}\\
    \end{align*}

    The CFG is explained below in the ``CFG Explanation'' section:\\

    CFG Explanation:\\
        P: The main program named as ``run'' which contains Statement S\\
        S: Consists of statements such as WHILE, IF, IFELSE, REPEAT. Can have multiple statements ``S S'' or action ``A''\\
        B: Perception H or not perception H that returns true or false\\
        H: Some boolean variables that provide the idea of the environment with true or false\\
        A: Action to be taken by the agent\\
        N: A positive integer that indicates the number of repetitions\\
        frontIsClear: Checks if the next cell towards the direction the agent is facing is inside the grid\\
        leftIsClear: Checks if the next cell to the left of the direction the agent is facing is inside the grid\\
        rightIsClear: Checks if the next cell to the right of the direction the agent is facing is inside the grid\\
        move: The agent moves towards its front\\
        turnLeft: The agent turns left\\
        turnRight: The agent turns right\\
        putMarker: The agent puts a marker in the current cell\\
        pickMarker: The agent picks a marker from the current cell\\
        ...: It is not part of the CFG. It has been used to indicate all positive numbers in between.\\

        To write a program from the given CFG, the following ``Program Writing Guidelines'' must be followed:\\
        
        Program Writing Guidelines:\\
        1. In the CFG, `infinity' means that the value of N can be up to infinity. So do not write any part of the program like ``R=infinity''\\
        2. DO NOT write ``...'' in the program, since it is not a part of the CFG\\
        3. The program must start with ``run''\\

        Now your tasks are the following 3:\\
        
        1. Read carefully about the details of the environment, the CFG and its explanation.\\
        2. Follow the CFG and the program writing guidelines and write a program of maximum 8 lines that will gain the maximum reward inside this given environment.\\
        3. Write the program inside <program></program> tag.
    
\end{tcolorbox}

\subsection{\textsc{Environment Details with More Information}} \label{sub:more_info}

\subsubsection*{StairClimber}
\begin{tcolorbox}[breakable, width=\textwidth, colback=white]
    Consider a 12x12 gridded 2D environment in Karel the Robot, where an agent has to reach a marker by climbing up along a stair. The grid contains a stair-like structure and the initial position of the agent is randomly assigned near the stair with the marker placed at the higher end, at the beginning of each episode.
\end{tcolorbox}

\subsubsection*{FourCorners}
\begin{tcolorbox}[breakable, width=\textwidth, colback=white]
    Consider a 12x12 gridded 2D environment in Karel the Robot, where an agent has to place one marker in each of the four corner cells of the grid. The grid is initially empty and the initial position of the agent is randomly assigned near the wall of the grid at the beginning of each episode.
\end{tcolorbox}

\subsubsection*{TopOff}
\begin{tcolorbox}[breakable, width=\textwidth, colback=white]
    Consider a 12x12 gridded 2D environment in Karel the Robot, where an agent has to place markers on top of other cells that already have markers, and then reach at the rightmost cell of the bottom row. The markers are initialized randomly at the bottom row. The position of the agent is fixed at the leftmost cell of the bottom row at the beginning of each episode.
\end{tcolorbox}

\subsubsection*{Maze}
\begin{tcolorbox}[breakable, width=\textwidth, colback=white]
    Consider a 12x12 gridded 2D environment in Karel the Robot, where an agent has to pick a marker following a path surrounded by walls. Some cells of the grid are filled with walls, collectively referred to as the maze. The agent, the marker and the maze are randomly placed at the beginning of each episode.
\end{tcolorbox}

\subsubsection*{CleanHouse}
\begin{tcolorbox}[breakable, width=\textwidth, colback=white]
    Consider a 22x14 gridded 2D environment in Karel the Robot, where an agent has to pick some markers that are placed randomly inide the grid. There are also some cells filled with obstacles. The position of the agent is fixed whereas, the markers are initialized at random cells at the beginning of each episode.
\end{tcolorbox}

\subsubsection*{Harvester}
\begin{tcolorbox}[breakable, width=\textwidth, colback=white]
    Consider an 8x8 gridded 2D environment in Karel the Robot, where an agent has to pick one marker from every cell of the grid. The grid is initially filled with markers and the initial position of the agent is randomly assigned at the beginning of each episode.
\end{tcolorbox}

\subsubsection*{DoorKey}
\begin{tcolorbox}[breakable, width=\textwidth, colback=white]
    Consider an 8x8 gridded 2D environment divided into two chambers by walls in Karel the Robot, where an agent has to pick a marker from the left chamber and put it over another marker placed at the right chamber. The agent cannot access the right chamber without picking the marker from the left chamber. The position of the agent, the marker of the left chamber and the marker of the right chamber are initialized randomly at the beginning of each episode.
\end{tcolorbox}

\subsubsection*{OneStroke}
\begin{tcolorbox}[breakable, width=\textwidth, colback=white]
    Consider an 8x8 gridded 2D environment in Karel the Robot, where an agent has to visit as many cells as possible in one attempt. Once the agent visits a cell, it will be filled with a wall. The position of the agent is initialized randomly at the beginning of each episode.
\end{tcolorbox}

\subsubsection*{Seeder}
\begin{tcolorbox}[breakable, width=\textwidth, colback=white]
    Consider an 8x8 gridded 2D environment in Karel the Robot, where an agent has to place one marker in every cell of the grid. The grid is initially empty and the initial position of the agent is randomly assigned at the beginning of each episode.
\end{tcolorbox}

\subsubsection*{Snake}
\begin{tcolorbox}[breakable, width=\textwidth, colback=white]
    Consider an 8x8 gridded 2D environment in Karel the Robot, where an agent has to pick a marker multiple times inside the grid. The position of the marker will be changed once the agent picks the marker. Each time the agent picks a marker, it will be attached to its body. For instance, if the agent picks 3 markers, it will have 3 markers added to its back. The agent has to pick as many markers as possible without hitting the markers attached to its body. The position of the agent and the marker are initialized randomly at the beginning of each episode.
\end{tcolorbox}

\subsection{\textsc{Environment Details with Less Information}} \label{sub:less_info}

\subsubsection*{StairClimber}
\begin{tcolorbox}[breakable, width=\textwidth, colback=white]
    Consider a 2D 12x12 grid with an agent placed randomly in any cell, which will interact within the grid. Each cell may contain a marker, be empty, or be blocked by a wall. Initially, for this particular problem, there is a stair-like structure and the agent is placed near the stair.

    The robot will receive different reward values depending on its interactions inside the grid. The interaction of the robot will be decided by a program written in the domain-specific language to be provided below as a context-free grammar (CFG). The goal is to generate a program that will maximize the sum of rewards the robot obtains by following that program.
\end{tcolorbox}

\subsubsection*{FourCorners}
\begin{tcolorbox}[breakable, width=\textwidth, colback=white]
    Consider a 2D 12x12 grid with an agent placed randomly in any cell, which will interact within the grid. Each cell may contain a marker, be empty, or be blocked by a wall. Initially, for this particular problem, no markers are present in any cell of the grid.

    The robot will receive different reward values depending on its interactions inside the grid. The interaction of the robot will be decided by a program written in the domain-specific language to be provided below as a context-free grammar (CFG). The goal is to generate a program that will maximize the sum of rewards the robot obtains by following that program.
\end{tcolorbox}

\subsubsection*{TopOff}
\begin{tcolorbox}[breakable, width=\textwidth, colback=white]
    Consider a 2D 12x12 grid with an agent placed at a fixed cell at the leftmost cell of the bottom row, which will interact within the grid. Each cell may contain a marker, be empty, or be blocked by a wall. Initially, for this particular problem, there are markers present in some cells at the bottom row of the grid.

    The robot will receive different reward values depending on its interactions inside the grid. The interaction of the robot will be decided by a program written in the domain-specific language to be provided below as a context-free grammar (CFG). The goal is to generate a program that will maximize the sum of rewards the robot obtains by following that program.
\end{tcolorbox}

\subsubsection*{Maze}
\begin{tcolorbox}[breakable, width=\textwidth, colback=white]
    Consider a 2D 12x12 grid with an agent placed randomly in any cell, which will interact within the grid. Each cell may contain a marker, be empty, or be blocked by a wall. Initially, for this particular problem, there are few walls and a marker inside the grid.

    The robot will receive different reward values depending on its interactions inside the grid. The interaction of the robot will be decided by a program written in the domain-specific language to be provided below as a context-free grammar (CFG). The goal is to generate a program that will maximize the sum of rewards the robot obtains by following that program.
\end{tcolorbox}

\subsubsection*{CleanHouse}
\begin{tcolorbox}[breakable, width=\textwidth, colback=white]
    Consider a 2D 22x14 grid with an agent placed at a fixed cell, which will interact within the grid. Each cell may contain a marker, be empty, or be blocked by a wall. Initially, for this particular problem, there are some markers and obstacles randomly placed inside the grid.

    The robot will receive different reward values depending on its interactions inside the grid. The interaction of the robot will be decided by a program written in the domain-specific language to be provided below as a context-free grammar (CFG). The goal is to generate a program that will maximize the sum of rewards the robot obtains by following that program.
\end{tcolorbox}

\subsubsection*{Harvester}
\begin{tcolorbox}[breakable, width=\textwidth, colback=white]
    Consider a 2D 8x8 grid with an agent placed randomly in any cell, which will interact within the grid. Each cell may contain a marker, be empty, or be blocked by a wall. Initially, for this particular problem, a marker is present in each cell of the grid.

    The robot will receive different reward values depending on its interactions inside the grid. The interaction of the robot will be decided by a program written in the domain-specific language to be provided below as a context-free grammar (CFG). The goal is to generate a program that will maximize the sum of rewards the robot obtains by following that program.
\end{tcolorbox}

\subsubsection*{DoorKey}
\begin{tcolorbox}[breakable, width=\textwidth, colback=white]
    Consider a 2D 8x8 grid divided into two chambers with an agent placed randomly in the left chamber, which will interact within the grid. Each cell may contain a marker, be empty, or be blocked by a wall. Initially, for this particular problem, there is one marker in each chamber of the grid.

    The robot will receive different reward values depending on its interactions inside the grid. The interaction of the robot will be decided by a program written in the domain-specific language to be provided below as a context-free grammar (CFG). The goal is to generate a program that will maximize the sum of rewards the robot obtains by following that program.
\end{tcolorbox}

\subsubsection*{OneStroke}
\begin{tcolorbox}[breakable, width=\textwidth, colback=white]
    Consider a 2D 8x8 grid with an agent placed randomly in any cell, which will interact within the grid. Each cell may contain a marker, be empty, or be blocked by a wall. Initially, for this particular problem, no markers are present in any cell of the grid. Cells that will be visited by the agent will be filled with obstacles.

    The robot will receive different reward values depending on its interactions inside the grid. The interaction of the robot will be decided by a program written in the domain-specific language to be provided below as a context-free grammar (CFG). The goal is to generate a program that will maximize the sum of rewards the robot obtains by following that program.
\end{tcolorbox}

\subsubsection*{Seeder}
\begin{tcolorbox}[breakable, width=\textwidth, colback=white]
    Consider a 2D 8x8 grid with an agent placed randomly in any cell, which will interact within the grid. Each cell may contain a marker, be empty, or be blocked by a wall. Initially, for this particular problem, no markers are present in any cell of the grid.

    The robot will receive different reward values depending on its interactions inside the grid. The interaction of the robot will be decided by a program written in the domain-specific language to be provided below as a context-free grammar (CFG). The goal is to generate a program that will maximize the sum of rewards the robot obtains by following that program.
\end{tcolorbox}

\subsubsection*{Snake}
\begin{tcolorbox}[breakable, width=\textwidth, colback=white]
    Consider a 2D 8x8 grid with an agent placed randomly in any cell, which will interact within the grid. Each cell may contain a marker, be empty, or be blocked by a wall. Initially, for this particular problem, there is one marker inside the grid. The position of the marker changes depending on a certain behaviour of the agent.

    The robot will receive different reward values depending on its interactions inside the grid. The interaction of the robot will be decided by a program written in the domain-specific language to be provided below as a context-free grammar (CFG). The goal is to generate a program that will maximize the sum of rewards the robot obtains by following that program.
\end{tcolorbox}

\section{\textsc{MicroRTS Maps}} \label{sec:mrts_maps}
\begin{figure}[h]
    \centering
    \includegraphics[width=0.37\linewidth]{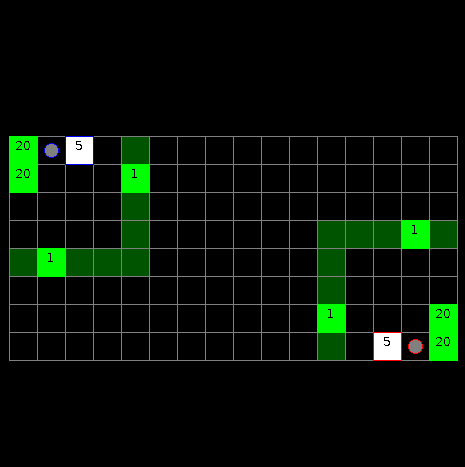}
    \includegraphics[width=0.37\linewidth]{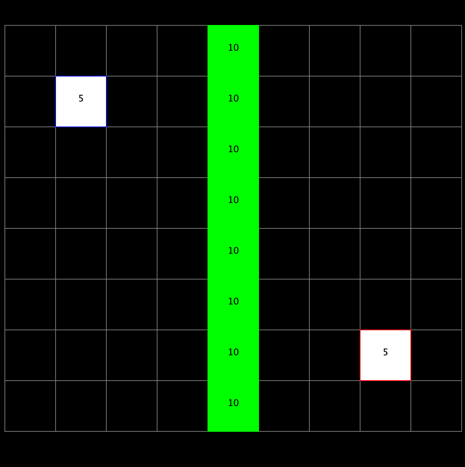}
    \includegraphics[width=0.37\linewidth]{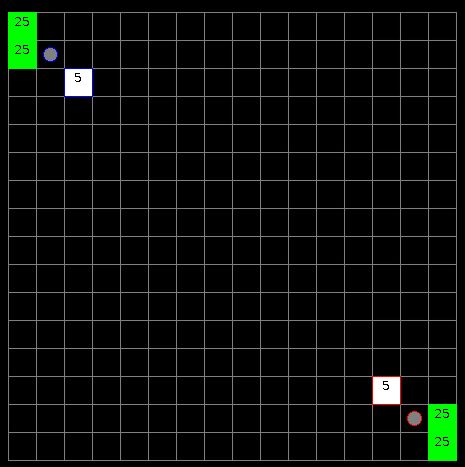}
    \includegraphics[width=0.37\linewidth]{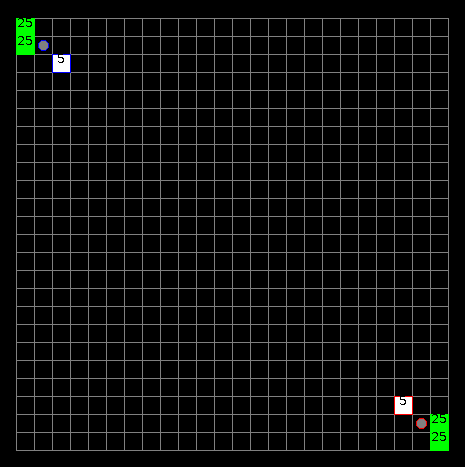}
    \includegraphics[width=0.37\linewidth]{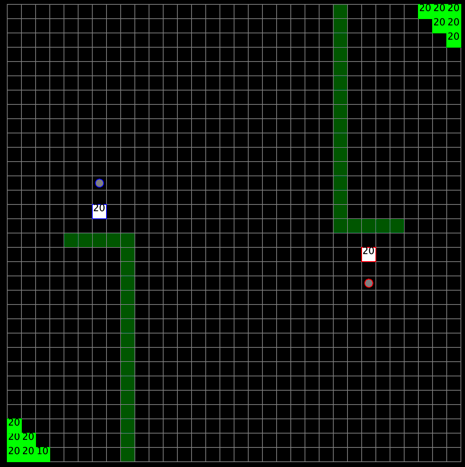}
    \includegraphics[width=0.37\linewidth]{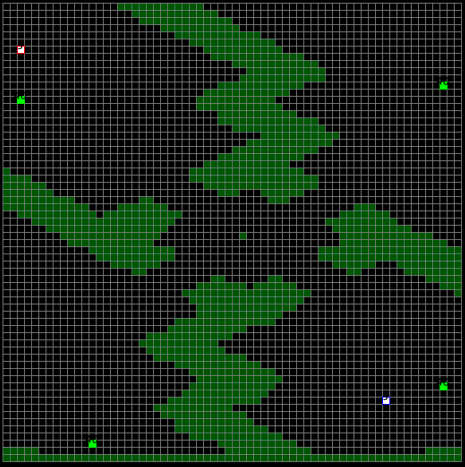}
    \caption{From left to right: LetMeOut (16$\times$8), NoWhereToRun (9$\times$8), BasesWorkers(16$\times$16), BasesWorkers(24$\times$24), BWDistantResources (32$\times$32), BloodBath (64$\times$64)}
    \label{fig:maps}
\end{figure}

Figure \ref{fig:maps} shows the three MicroRTS maps used in our experiments. In these maps, the geometric shapes with blue borders represent the units of one player, while those bounded by red borders represent the units of the other player. Neutral objects do not have colored borders (e.g., resources in light green and walls in dark green).

\clearpage

\section{\textsc{MicroRTS Prompts}} \label{sec:mrts_prompts}
In this section, we present the prompt used to obtain one program that encodes a policy for \textsc{MicroRTS}. This is the prompt that we use multiple times to obtain a set of programmatic policies. Section~\ref{sub:mrts_full_prompt} shows the complete prompt for the `NoWhereToRun (9x8)' map, as a complete example. Section~\ref{sub:mrts_env} shows the description of the environment used in the prompts of each map.

\subsection{\textsc{Complete Prompt for the NoWhereToRun (9x8) Map}} \label{sub:mrts_full_prompt}
The prompt for obtaining a program that encodes policies for the `NoWhereToRun (9x8)' map is shown below. Note that the first paragraph of the prompt mentions the details of the environment for a given map. Only this paragraph describing the environment is updated to obtain the program for each separate map. We use a subset of the MicroLanguage, where conditionals are removed. Our goal was to have a language that would be easier for the foundation model to use. 

\begin{tcolorbox}[breakable, width=\textwidth, colback=white]
Consider a 9x8 gridded map of microRTS, a real-time strategy game. Consider this map as a 2-dimensional array with the following structure:
\begin{itemize}
    \item[--] There are a total of 8 neutral resource cells situated along the central column of the map, dividing the map into two parts. Each resource cell contains 10 units of resources.
    \item[--] The base B1 of player 1 is located at index (1,1), which is located on the left side of the map.
    \item[--] The base B2 of player 2 is located at index (7,6), which is located on the right side of the map.
    \item[--] Each player controls one base, which initially has 5 units of resources.
    \item[--] The only unit a player controls at the beginning of the game is the base.
\end{itemize}

Consider this Context-Free Grammar (CFG) describing a programming language for writing programs encoding strategies of microRTS. The CFG is shown in  the $<CFG></CFG>$ tag bellow:

\begin{align*}
\text{$<CFG>$} \\
S &\to S S \mid \text{for(Unit u) S} \mid C \mid \lambda \\
C &\to u.\text{build}(T, D, N) \mid u.\text{train}(T, D, N) \mid u.\text{moveToUnit}(T_p, O_p) \\
&\quad \mid u.\text{attack}(O_p) \mid u.\text{harvest}(N) \mid u.\text{attackIfInRange()} \mid u.\text{moveAway()}\\
T &\to \text{Base} \mid \text{Barracks} \mid \text{Ranged} \mid \text{Heavy} \\
&\quad \mid \text{Light} \mid \text{Worker} \\
N &\to 0 \mid 1 \mid 2 \mid 3 \mid 4 \mid 5 \mid 6 \mid 7 \mid 8 \mid 9 \\
&\quad \mid 10 \mid 15 \mid 20 \mid 25 \mid 50 \mid 100\\
D &\to \text{EnemyDir} \mid \text{Up} \mid \text{Down} \mid \text{Right} \mid \text{Left}\\
O_p &\to \text{Strongest} \mid \text{Weakest} \mid \text{Closest} \mid \text{Farthest} \\
&\quad \mid \text{LessHealthy} \mid \text{MostHealthy} \mid \text{Random}\\
T_p &\to \text{Ally} \mid \text{Enemy}\\
\text{$</CFG>$}
\end{align*}

This language allows nested loops. It contains several command-oriented functions (C).

The Command functions (`C' in the CFG) are described below:
\begin{enumerate}
\item  u.build(T, D, N): Builds N units of type T on a cell located in the D direction of the unit. The u.build function is used to build Barracks and Base.
\item  u.train(T, D, N): Trains N units of type T on a cell located in the D direction of the structure responsible for training them. For example, the instruction u.train(Heavy, Down, 1) will allow the agent to train at most 1 heavy unit in the down direction of the Barrack, while the instruction u.train(Heavy, EnemyDir, 20) will allow to train at most 20 towards the direction of the opponent. The number used in the function calls could play a big role in the strategy the program encodes. The u.train function is used to train Worker, Ranged, Light, and Heavy units.
\item  u.moveToUnit(T\_p, O\_p): Commands a unit to move towards the player T\_p following a criterion O\_p.
\item  u.attack(O\_p): Sends N Worker units to harvest resources.
\item  u.harvest(N): Sends N Worker units to harvest resources. For example, u.harvest(5) will send 5 workers to harvest resources.
\item  u.attackIfInRange(): Commands a unit to stay idle and attack if an opponent unit comes within its attack range.
\item  u.moveAway(): Commands a unit to move in the opposite direction of the player's base.
\end{enumerate}

`T' represents the types a unit can assume. 

`D' represents the directions available used in action functions.

`O\_p' is a set of criteria to select an opponent unit based on their current state.

`T\_p' represents the set of target players.
            
`N' is the number of units that can be any integers from 0 to 10, or 15, or 20, or 25, or 50, or 100.\\

The following 4 are some guidelines for writing the playing strategy:
\begin{enumerate}

\item There is NO NEED TO write classes, or initiate objects such as Unit, Worker, etc. There is also NO NEED TO write comments.
\item Use curly braces like C/C++/Java while writing any `for' block. Start the curly braces in the same line of the block.
\item A strategy must be written inside one or multiple `for' blocks.
\item This language does not have if-statements.
\end{enumerate}

Now your tasks are the following 8:

\begin{enumerate}
    \item Understand the command (C) functions from above and try to relate them in the context of playing strategies for a real-time strategy game.
    \item Write a program in the microRTS language encoding a very strong game-playing strategy for the map described above. You must follow the guidelines of writing the playing strategy while writing your program.
    \item You must not use any symbols (for example \&\&, $\parallel$, etc.) that the CFG does not accept. You have to strictly follow the CFG while writing the program.
    \item Look carefully, the methods of non-terminal symbols C have prefixes 'u.' in the examples since they are methods of the object 'Unit u'. You should follow the patterns of the examples.
    \item Write only the pseudocode inside `$<strategy></strategy>$' tag.
    \item Do not write unnecessary symbols of the CFG such as, `$S \to$', `$\to$', etc.
    \item Check the program and ensure it does not violate the rules of the CFG or the guidelines for writing the strategy.
    \item The for loops in this language iterate over all units and the instructions inside the for loops attempt to assign actions to each of these units. That is why having nested for loops allows for a prioritization scheme. The innermost for-loops will contain the actions with the highest priority.  Effective programs usually have nested for-loops.
\end{enumerate}

\end{tcolorbox}

\subsection{\textsc{Environment Details}} \label{sub:mrts_env}

\subsubsection*{NoWhereToRun (9x8)}
\begin{tcolorbox}[breakable, width=\textwidth, colback=white]
    Consider a 9x8 gridded map of microRTS, a real-time strategy game. Consider this map as a 2-dimensional array with the following structure:
    \begin{itemize}
        \item[--] There are a total of 8 neutral resource cells situated along the central column of the map, dividing the map into two parts. Each resource cell contains 10 units of resources.
        \item[--] The base B1 of player 1 is located at index (1,1), which is located on the left side of the map.
        \item[--] The base B2 of player 2 is located at index (7,6), which is located on the right side of the map.
        \item[--] Each player controls one base, which initially has 5 units of resources.
        \item[--] The only unit a player controls at the beginning of the game is the base.
    \end{itemize}
\end{tcolorbox}

\subsubsection*{DoubleGame (24x24)}
\begin{tcolorbox}[breakable, width=\textwidth, colback=white]
    Consider a 24x24 gridded map of microRTS, a real-time strategy game. Consider this map as a 2-dimensional array with the following structure:
    \begin{itemize}
        \item[--] There is a wall in the middle of the map consisting of two columns that has a small passage of 4 cells. The small passage consists of 4 resource cells each having only 1 resource.
        \item[--] There are 28 resource cells at the top-left, top-right, bottom-left and bottom-right corners of the map respectively where each of them contains 10 units of resources.
        \item[--] The bases of player 1 are located at indices (3,2) and (20,2), located on both sides of the wall.
        \item[--] The bases of player 2 are located at indices (20,21) and (3,21), also located on both sides of the wall.
        \item[--] Each player controls two bases, which initially have 5 units of resources each.
        \item[--] There are 2 workers beside each base. So a total of 4 workers for each of the players.
    \end{itemize}
\end{tcolorbox}

\subsubsection*{BWDistantResources (32x32)}
\begin{tcolorbox}[breakable, width=\textwidth, colback=white]
    Consider a 32x32 map of microRTS, a real-time strategy game. Consider this map as a 2-dimensional array with the following structure:
    \begin{itemize}
        \item[--] There are two L-shaped obstacles on the map, each with a passage of 4 cells located at the middle of left and right sides.
        \item[--] There are a total of 12 neutral resource cells R located at the top-right and bottom-left corners of the map. Each resource center contains 20 units of resources.
        \item[--] The base B1 of player 1 is located at index (6,14), which is located on the left side of the map.
        \item[--] The base B2 of player 2 is located at index (25,17), which is located on the right side of the map.
        \item[--] Each player controls one Base, which initially has 20 units of resources.
        \item[--] There is one worker for each player besides their bases.
    \end{itemize}
\end{tcolorbox}

\subsubsection*{BloodBath (64x64)}
\begin{tcolorbox}[breakable, width=\textwidth, colback=white]
    Consider a 64x64 gridded map of microRTS, a real-time strategy game. Consider this map as a 2-dimensional array with the following structure:
    \begin{itemize}
        \item[--] There are total 4 neutral resource cells situated close to the top-left, top-right, bottom-left and bottom-right sides of the map respectively. Each resource cell contains 40 units of resources.
        \item[--] There are obstacles in between each of the 4 resource centers.
        \item[--] The base B1 of player 1 is located at index (53, 55), which is located on the bottom-right side of the map.
        \item[--] The base B2 of player 2 is located at index (2, 6), which is located on the top-left side of the map.
        \item[--] Each player controls one Base each, which initially has 5 units of resources.
        \item[--] There is no worker for both player 1 and 2 in the initial map setup.
        \item[--] The only unit a player controls at the beginning of the game is the Base.
    \end{itemize}
\end{tcolorbox}

\section{\textsc{Self-Play Learning Algorithms}} \label{sec:local_learner}


Self-play algorithms attempt to approximate an optimal policy for two-player games. 
Iterated Best Response (IBR)~\cite{lanctot2017unified} is perhaps the simplest self-play algorithm that we could use. IBR starts with an arbitrary policy $\pi_0$ in $\llbracket D \rrbracket$ for one of the players and approximates a best response $\pi_1$ to $\pi_0$ for the other player. Then it approximates a best response $\pi_2$ to $\pi_1$ for the first player. This process is repeated a number of times $m$, which is normally determined by a computational budget. The last resulting policy $\pi_m$ is returned as IBR's approximate optimal policy for the game.

The self-play process IBR follows generates a sequence of policies for the players, but IBR only considers the latest policy while computing a best response. Other algorithms, such as Fictitious Play (FP)~\cite{browniterative}, compute best responses to a policy that mixes the best responses computed in all previous iterations. The use of more policies allows the method to find optimal policies even in games with cyclic dynamics such as Rock, Paper, and Scissors.

The algorithm we use in our experiments, Local Learner (2L)~\cite{locallearner}, also considers multiple policies in its self-play loop. This allows 2L to use more information than IBR. However, instead of including all policies seen in the process, like FP does, it selects a subset of the policies seen in the process. Using all policies can be computationally wasteful, as many of the policies are ``redundant''. 2L selects which policies to use based on the information collected during the search for programmatic policies. We refer the reader to the work of \cite{locallearner} for a detailed explanation of 2L.

\end{document}